\documentclass[letterpaper, 10 pt, conference]{ieeeconf}
\IEEEoverridecommandlockouts
\usepackage[T1]{fontenc}    % use 8-bit T1 fonts
\usepackage{hyperref}       % hyperlinks
\usepackage{url}            % simple URL typesetting

\usepackage[utf8]{inputenc} % allow utf-8 input
\usepackage{subcaption} % 导入 subcaption 宏包

\usepackage[table]{xcolor} %gxd
\usepackage{xcolor}
\usepackage{amsmath}
\usepackage{amssymb}
\usepackage{multirow}
\usepackage{pifont}
\usepackage{pgffor}      % 用于实现循环
\usepackage{makecell}

\usepackage{booktabs}       % professional-quality tables
\usepackage{amsfonts}    % blackboard math symbols
\usepackage{nicefrac}    % compact symbols for 1/2, etc.
\usepackage{microtype}      % microtypography
\usepackage{adjustbox} %用于figure5垂直注释居中

\definecolor{BlueGreen}{HTML}{008080}
\definecolor{OliveGreen}{RGB}{128, 128, 0}

\definecolor{mydarkdarkgreen}{RGB}{93, 150, 74} % 用于字体颜色标记 & 用于打勾
\definecolor{mydarkgreen}{RGB}{216, 233, 199}  % 段首颜色标记
\definecolor{mylightgreen}{RGB}{245, 249, 241}

\definecolor{RedOrange}{RGB}{255, 165, 0}

\definecolor{lightgray}{gray}{.9}
\newcolumntype{I}{!{\vrule width 1pt}}
\makeatletter
\newcommand{\thickhline}{%
    \noalign {\ifnum 0=`}\fi \hrule height 1pt
    \futurelet \reserved@a \@xhline
}
\makeatother

\begin{document}

%%%%%%%%% TITLE
\title{StereoCarla: A High-Fidelity Driving Dataset for Generalizable Stereo }
% diverse baselines matter for synthetic stereo training

\author{
\vspace{-5mm}
 \\
    \textbf{Xianda Guo}\textsuperscript{1,$^*$},
    \textbf{Chenming Zhang}\textsuperscript{2,3,$^*$},
    \textbf{Ruilin Wang}\textsuperscript{4},
    \textbf{Youmin Zhang}\textsuperscript{5},
    \textbf{Wenzhao Zheng}\textsuperscript{6},\\
    \textbf{Matteo Poggi}\textsuperscript{5},
    \textbf{Hao Zhao}\textsuperscript{7},
    \textbf{Qin Zou}\textsuperscript{1,$^\dagger$},
    \textbf{Long Chen}\textsuperscript{4,2,3,$^\dagger$} \\
    \textsuperscript{1} School of Computer Science, Wuhan University~~~~\textsuperscript{2} IAIR, Xi'an Jiaotong University
    ~~~~\textsuperscript{3} Waytous \\~~~~\textsuperscript{4} CASIA~~~~\textsuperscript{5}University of Bologna~~~~\textsuperscript{6} University of California, Berkeley
\textsuperscript{7} AIR, Tsinghua University
\\
\texttt{xianda\_guo@163.com}
\vspace{-5mm}
}

\maketitle

\renewcommand{\thefootnote}{\fnsymbol{footnote}}
\footnotetext[1]{These authors contributed equally to this work.}
\footnotetext[2]{Corresponding authors}

%%%%%%%%% ABSTRACT
\begin{abstract}

Stereo matching plays a crucial role in enabling depth perception for autonomous driving and robotics. While recent years have witnessed remarkable progress in stereo matching algorithms, largely driven by learning-based methods and synthetic datasets, the generalization performance of these models remains constrained by the limited diversity of existing training data. 
To address these challenges, we present StereoCarla, a high-fidelity synthetic stereo dataset specifically designed for autonomous driving scenarios. Built on the CARLA simulator, StereoCarla incorporates a wide range of camera configurations—including diverse baselines, viewpoints, and sensor placements—as well as varied environmental conditions such as lighting changes, weather effects, and road geometries. 
We conduct comprehensive cross-domain experiments across four standard evaluation datasets (KITTI2012, KITTI2015, Middlebury, ETH3D) and demonstrate that models trained on StereoCarla outperform those trained on 11 existing stereo datasets in terms of generalization accuracy across multiple benchmarks.
Furthermore, when integrated into multi-dataset training, StereoCarla contributes substantial improvements to generalization accuracy, highlighting its compatibility and scalability.
This dataset provides a valuable benchmark for developing and evaluating stereo algorithms under realistic, diverse, and controllable settings, facilitating more robust depth perception systems for autonomous vehicles.
Code can be available at \url{https://github.com/XiandaGuo/OpenStereo} and
data can be available at \url{https://xiandaguo.net/StereoCarla}.

\end{abstract}
%%%%%%%%% BODY TEXT
\section{Introduction}

Robust depth perception is fundamental to autonomous driving and robotics~\cite{guo2023simple,zhang2023completionformer, duan2023diffusiondepth}, enabling vehicles to accurately interpret their surroundings and make safe navigation decisions. Among various depth sensing technologies, stereo vision stands out as a passive, cost-effective, and widely adopted method, which estimates dense disparity maps from rectified stereo image pairs. Unlike LiDAR-based approaches that require expensive sensors and post-processing, stereo vision systems leverage affordable cameras and offer dense depth outputs with relatively low latency. These advantages make stereo vision an attractive solution for practical deployment in autonomous driving platforms.

Recent years have witnessed remarkable advances in stereo matching, driven by deep learning architectures such as PSMNet~\cite{psmnet2018}, RAFT-Stereo~\cite{raftstereo}, and NMRF-Stereo~\cite{NMRFStereo}, together with the availability of large-scale datasets~\cite{sceneflow,kitti2012,kitti2015,yang2019drivingstereo,karaev2023dynamicstereo}. While these developments have significantly improved benchmark performance, the generalization ability of current models remains a persistent challenge. Models trained on a specific dataset often degrade substantially when deployed in unseen environments, limiting their robustness in real-world driving.

\begin{figure}[t]
    \centering
    \includegraphics[width=1\linewidth]{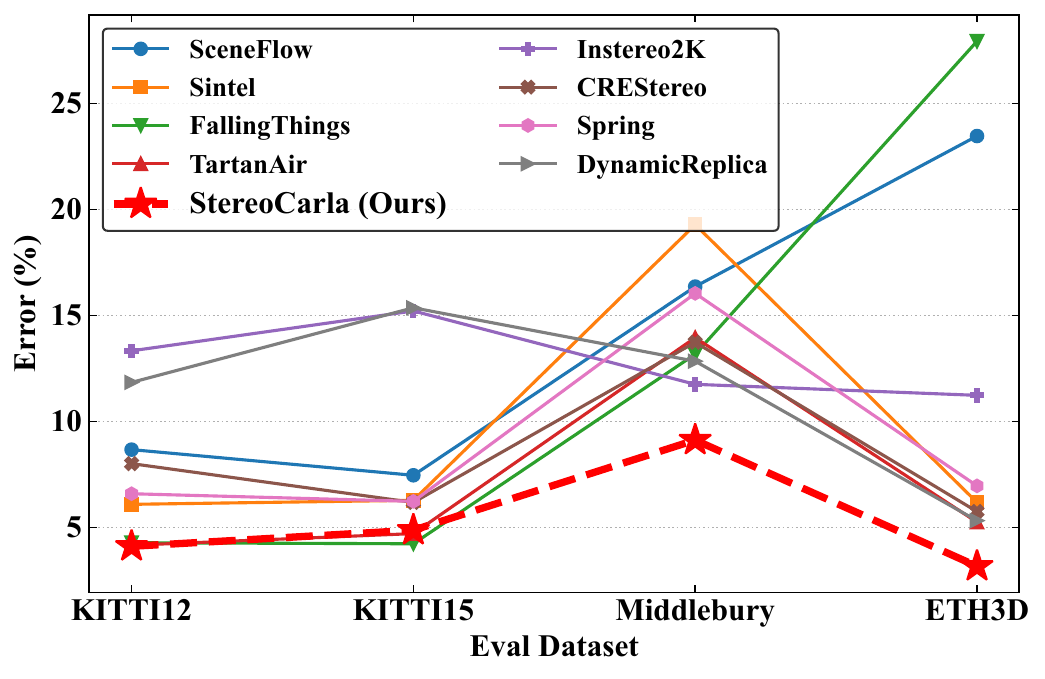}
    \caption{\textbf{Zero-shot performance of stereo models trained on different datasets.} Models trained on our StereoCarla datasets consistently achieve superior generalization, outperforming those trained on other existing stereo datasets.}
    \label{fig:teaser}
\end{figure}

Despite these advancements, the generalization capability of current stereo models remains limited, largely due to the scarcity and homogeneity of existing datasets. Most publicly available stereo datasets either focus on specific scenarios, such as urban driving scenes with fixed camera setups (e.g., KITTI~\cite{kitti2012,kitti2015}, DrivingStereo~\cite{yang2019drivingstereo}), or synthetic indoor scenes lacking real-world complexities (e.g., SceneFlow~\cite{sceneflow}). Others, such as VirtualKITTI~\cite{cabon2020virtual} and FallingThings~\cite{tremblay2018falling}, are constrained in visual fidelity or scenario realism, and often fail to provide the level of variability required for training general-purpose stereo models. As a result, existing datasets fail to sufficiently capture key factors in real-world driving, including variations in camera baselines, viewpoints, motion dynamics, weather conditions, lighting changes, and environmental textures. This limitation poses a significant obstacle to developing robust stereo algorithms capable of generalizing well to the highly dynamic and complex scenarios encountered in real-world autonomous driving.

To bridge this critical gap, we propose \textit{StereoCarla}, a novel high-fidelity synthetic stereo dataset specifically designed for generalizable stereo. Leveraging the powerful CARLA simulator, StereoCarla offers: (1) \textbf{Extensive camera diversity} — variable baselines, sensor heights, and pitch/roll angles. (2) \textbf{Rich environmental conditions} — diverse weather. % (3) \textbf{Dynamic scenarios} — realistic traffic agents, and motion patterns.
This comprehensive design allows systematic evaluation and training of stereo matching algorithms under realistic and highly controlled conditions, while avoiding the annotation costs and constraints of real-world data collection.

To validate the effectiveness of StereoCarla, we conduct extensive experiments on four challenging public benchmarks: KITTI 2012~\cite{kitti2012}, KITTI 2015~\cite{kitti2015}, Middlebury~\cite{middlebury}, and ETH3D~\cite{eth3d}. As shown in Fig.~\ref{fig:teaser}, models trained on StereoCarla consistently outperform those trained on other existing stereo datasets in terms of cross-domain generalization. Notably, although StereoCarla is collected from autonomous driving scenarios within the CARLA simulator, it still achieves superior zero-shot performance on the Middlebury and ETH3D datasets.
Furthermore, when used as a foundational component in multi-dataset training, StereoCarla continues to deliver substantial performance gains, reinforcing its value as a core dataset for stereo vision research. We believe StereoCarla provides the community with a scalable, diverse, and high-quality resource to drive the development of more generalizable and robust stereo matching systems.

\begin{table*}[t]
\caption{\textbf{Comparison of available stereo datasets.} In/out/Dy/W/Acc./Divers. refers to Indoor/Outdoor/Dynamic/Weather/Accuracy/Diversity. FL refers to focal length. Range refers to disparity range. Ave./Med. refers to the average/median of disparity.}
\label{tab:datasets_summary}
\centering
\scriptsize
\setlength\tabcolsep{8pt}
\renewcommand\arraystretch{1.1}
\resizebox{\linewidth}{!}{
\begin{tabular}{l|c|cccccc|cc|c|c|c|c|c|c}
\thickhline 
% \rowcolor{mydarkgreen} 
\rowcolor{gray!20}
\textbf{Dataset} & \textbf{Year} & In & Out & Dy & Video & Dense & W & Acc. & Divers. & Type & Resolution & Baseline & FL & Range & Ave./Med. \\
\hline\hline
KITTI12~\cite{kitti2012} & CVPR12 & \ding{55} & \checkmark & \checkmark & \checkmark & \ding{55} & \ding{55} & Med & Low & LiDAR & 1242×375 & 0.54m & 720px & 4–232 & 40.1/38 \\

\rowcolor{gray!10}
Sintel~\cite{Sintel} & ECCV12 & \checkmark & \checkmark & \checkmark & \checkmark & \checkmark & \ding{55} & High & Med & Syn & 1024×436 & 0.1m & - & 0–972 & 66.5/25 \\
KITTI15~\cite{kitti2015} & CVPR15 & \ding{55} & \checkmark & \checkmark & \checkmark & \ding{55} & \ding{55} & Med & Low & LiDAR & 1242×375 & 0.54m & 520px & 4–230 & 35.2/33 \\

\rowcolor{gray!10}
Middlebury~\cite{middlebury} & GCPR14 & \checkmark & \ding{55} & \ding{55} & \ding{55} & \ding{55} & \ding{55} & High & Low & LiDAR & ~6Mpx & 140–400mm & 1100–3600px & 15–323 & 72.5/63 \\
SceneFlow~\cite{sceneflow} & CVPR16 & \checkmark & \checkmark & \checkmark & \checkmark & \checkmark & \ding{55} & High & High & Syn & 960×540 & 0.54m & - & 0–10501 & 53.9/36 \\

\rowcolor{gray!10}
ETH3D~\cite{eth3d} & CVPR17 & \checkmark & \checkmark & \ding{55} & \checkmark & \ding{55} & \ding{55} & High & Low & LiDAR & ~0.3Mpx & 59.5–60.4mm & 529–712px & 0–62 & 13.7/10 \\
FallingThings~\cite{tremblay2018falling} & CVPRW18 & \checkmark & \checkmark & \ding{55} & \ding{55} & \checkmark & \ding{55} & High & Low & Syn & 960×540 & 6cm & 768.2px & 7–461 & 35.2/34 \\

\rowcolor{gray!10}
Argoverse~\cite{chang2019argoverse} & CVPR19 & \ding{55} & \checkmark & \checkmark & \checkmark & \ding{55} & \ding{55} & Low & Low & LiDAR & 2056×2464 & – & – & 0–256 & 69.1/59 \\
VKITTI2~\cite{cabon2020virtual} & ArXiv20 & \ding{55} & \checkmark & \checkmark & \checkmark & \checkmark & \ding{55} & High & Mid & Syn & 1242×375 & 0.54m & 725px & 0–411 & 30.1/25 \\

\rowcolor{gray!10}
TartanAir~\cite{wang2020tartanair} & IROS20 & \checkmark & \checkmark & \checkmark & \checkmark & \checkmark & \ding{55} & High & High & Syn & 640×480 & – & – & 0–499 & 21.0/13 \\
InStereo2K~\cite{bao2020instereo2k} & SCIS20 & \checkmark & \ding{55} & \ding{55} & \ding{55} & \ding{55} & \ding{55} & Low & Low & LiDAR & 1080×860 & 10cm & 8mm & 0–328 & 78.4/74 \\

\rowcolor{gray!10}
UnrealStereo4K~\cite{tosi2021smd} & CVPR21 & \checkmark & \checkmark & \ding{55} & \ding{55} & \checkmark & \ding{55} & High & High & Syn & 3840×2160 & 0.2m/0.5m & – & 0–1515 & 175.3/135 \\
CREStereo~\cite{Crestereo} & CVPR22 & \checkmark & \checkmark & \ding{55} & \ding{55} & \checkmark & \ding{55} & High & High & Syn & 1920×1080 & – & – & 0–2048 & 15.2/8 \\

\rowcolor{gray!10}
Spring~\cite{mehl2023spring} & CVPR23 & \ding{55} & \checkmark & \checkmark & \checkmark & \checkmark & \ding{55} & High & Low & Syn & 1920×1080 & 6.5cm & – & 0–554 & 38.1/19 \\
DynamicReplica~\cite{karaev2023dynamicstereo} & CVPR23 & \checkmark & \ding{55} & \checkmark & \checkmark & \checkmark & \ding{55} & High & Low & Syn & 1280×720 & 4–30cm & – & 3–656 & 62.7/48 \\
% \hline
\rowcolor{gray!10}
FoundationStereo~\cite{wen2025foundationstereo} & CVPR25 & \checkmark& \checkmark & \ding{55} & \ding{55} & \checkmark & \ding{55} & High & High & Syn & 1280×720 &- &- &1-700& 52.5/39\\

\rowcolor{cyan!10}
\textbf{StereoCarla (Ours)} & – & \ding{55} & \checkmark & \checkmark & \checkmark & \checkmark & \checkmark & High & High & Syn & 1600×900 & 10–300cm & 1385.6px & 0–3318 & 80.8/31 \\

\hline\hline

\thickhline
\end{tabular}}
\vspace{-5pt}
\end{table*}

Our contributions can be summarized as follows:
\begin{itemize} 
\item We build a new synthetic dataset to better generalize across different scenarios and enhance the performance, called StereoCarla. Compared to existing datasets, it features unique view angles, baselines, and weather conditions.

\item We systematically demonstrate through extensive experiments that stereo models trained on StereoCarla achieve superior generalization accuracy and robustness when compared to models trained on 11 widely-used stereo datasets, validating the effectiveness and practical utility of our dataset.

\item We provide comprehensive benchmarks and detailed evaluations on StereoCarla to facilitate future research. By making the dataset and evaluation framework publicly available, we aim to accelerate the development of robust and generalizable stereo matching methods.

\end{itemize}

\section{Related Work}

\subsection{Stereo Datasets}

As shown in Table~\ref{tab:datasets_summary}, recent years have witnessed significant growth in the number of publicly available datasets, varying substantially in scale, realism, and complexity.

Synthetic stereo datasets, such as SceneFlow~\cite{sceneflow}, Sintel~\cite{Sintel}, and FallingThings~\cite{tremblay2018falling}, provide dense ground-truth annotations, facilitating training and pre-training of stereo models. SceneFlow, composed of indoor and outdoor scenarios, has become a cornerstone in stereo vision research due to its large-scale and accurate ground-truth disparities. FallingThings further supports research in 3D object detection by providing highly detailed synthetic environments. Additionally, UnrealStereo4K~\cite{tosi2021smd} leverages advanced graphics techniques to deliver ultra-high-resolution stereo pairs, enhancing model precision under challenging visual conditions. Tartanair~\cite{wang2020tartanair} emphasizes diverse synthetic environments specifically for visual SLAM, contributing significantly to stereo vision research.

Real-world stereo datasets, including CREStereo~\cite{Crestereo}, InStereo2K~\cite{bao2020instereo2k}, and Spring~\cite{mehl2023spring}, offer critical diversity for training robust stereo algorithms. CREStereo introduces variations in lighting and color, focusing on challenging real-world conditions. InStereo2K provides a collection of indoor stereo images with high-quality, semi-dense disparities, addressing indoor scene complexities. Spring further expands real-world data availability with high-resolution stereo pairs across diverse scene compositions.

Datasets designed explicitly for autonomous driving scenarios, such as KITTI~\cite{kitti2012,kitti2015}, VirtualKITTI2~\cite{cabon2020virtual}, DrivingStereo~\cite{yang2019drivingstereo}, and DynamicStereo~\cite{karaev2023dynamicstereo}, have greatly impacted the evaluation standards for stereo methods. KITTI datasets, acquired through real-world driving sessions, have become standard benchmarks due to their precise LiDAR annotations. DrivingStereo offers extensive stereo image collections from diverse driving environments, using advanced annotation techniques involving LiDAR and deep learning. VirtualKITTI2, closely mimicking KITTI's real-world data via the Unity engine, further bridges synthetic-real gaps. DynamicStereo specifically addresses scenarios involving non-rigid objects, significantly enhancing stereo matching robustness.
Additionally, Middlebury~\cite{middlebury} and ETH3D~\cite{eth3d} datasets, capturing both indoor and outdoor scenes with structured-light systems and laser scanners, respectively, serve as standard benchmarks for evaluating the cross-domain generalization performance of stereo methods.

Despite these significant contributions, existing datasets frequently exhibit limited variations in camera baselines, viewpoints, and weather conditions, restricting stereo models' generalization capability across diverse autonomous driving scenarios. Motivated by these limitations, our proposed dataset, StereoCarla, enriches the diversity and realism in stereo data generation, introducing controllable variations that surpass existing benchmarks, aiming to significantly improve model robustness and generalization in practical autonomous driving applications.

\begin{figure*}[t]
    \centering
    % 第一行图片
    \begin{subfigure}[b]{0.15\textwidth}
        \centering
        \includegraphics[width=\textwidth]{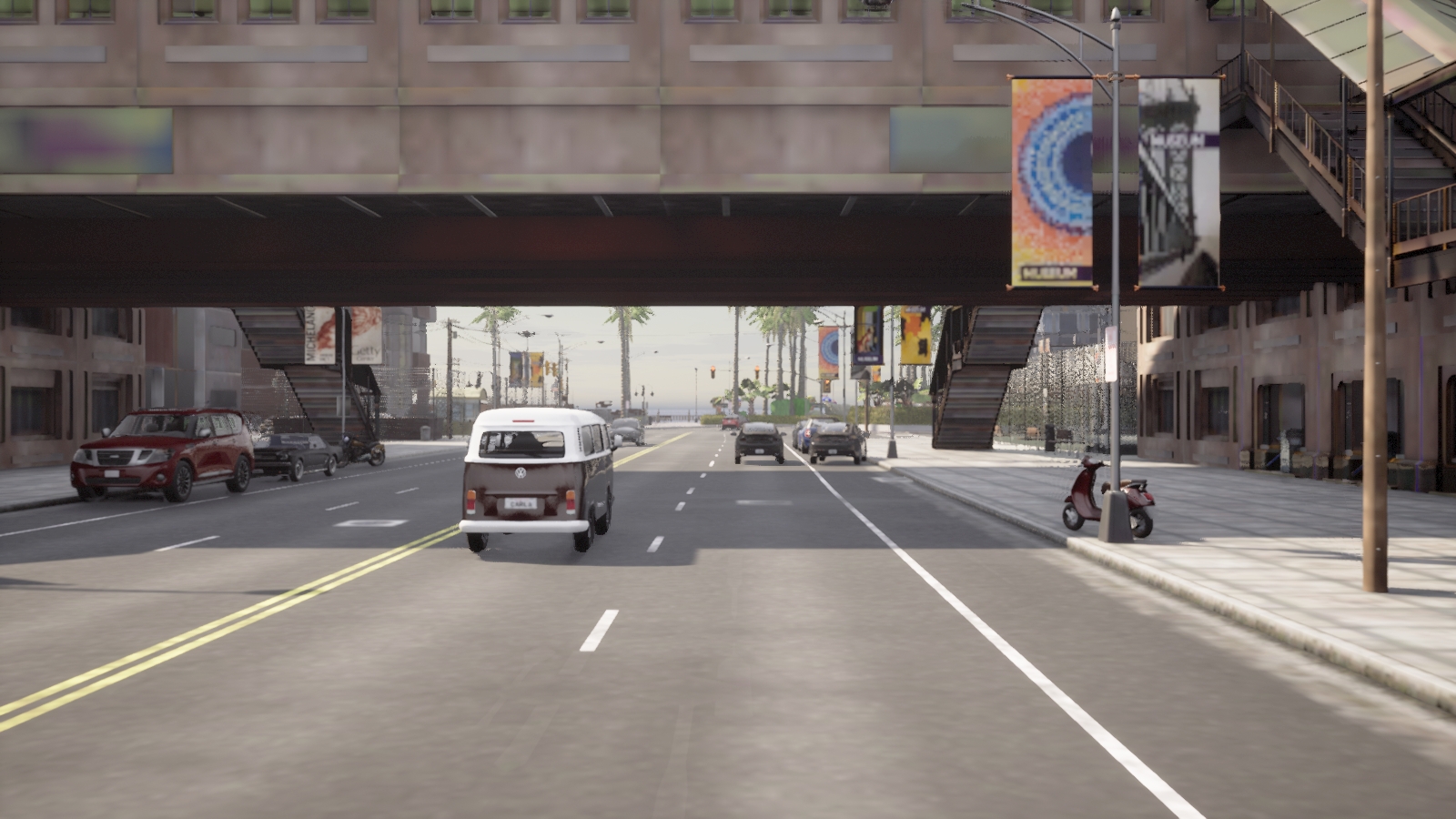}
        \caption{Left}
    \end{subfigure}
    \begin{subfigure}[b]{0.15\textwidth}
        \centering
        \includegraphics[width=\textwidth]{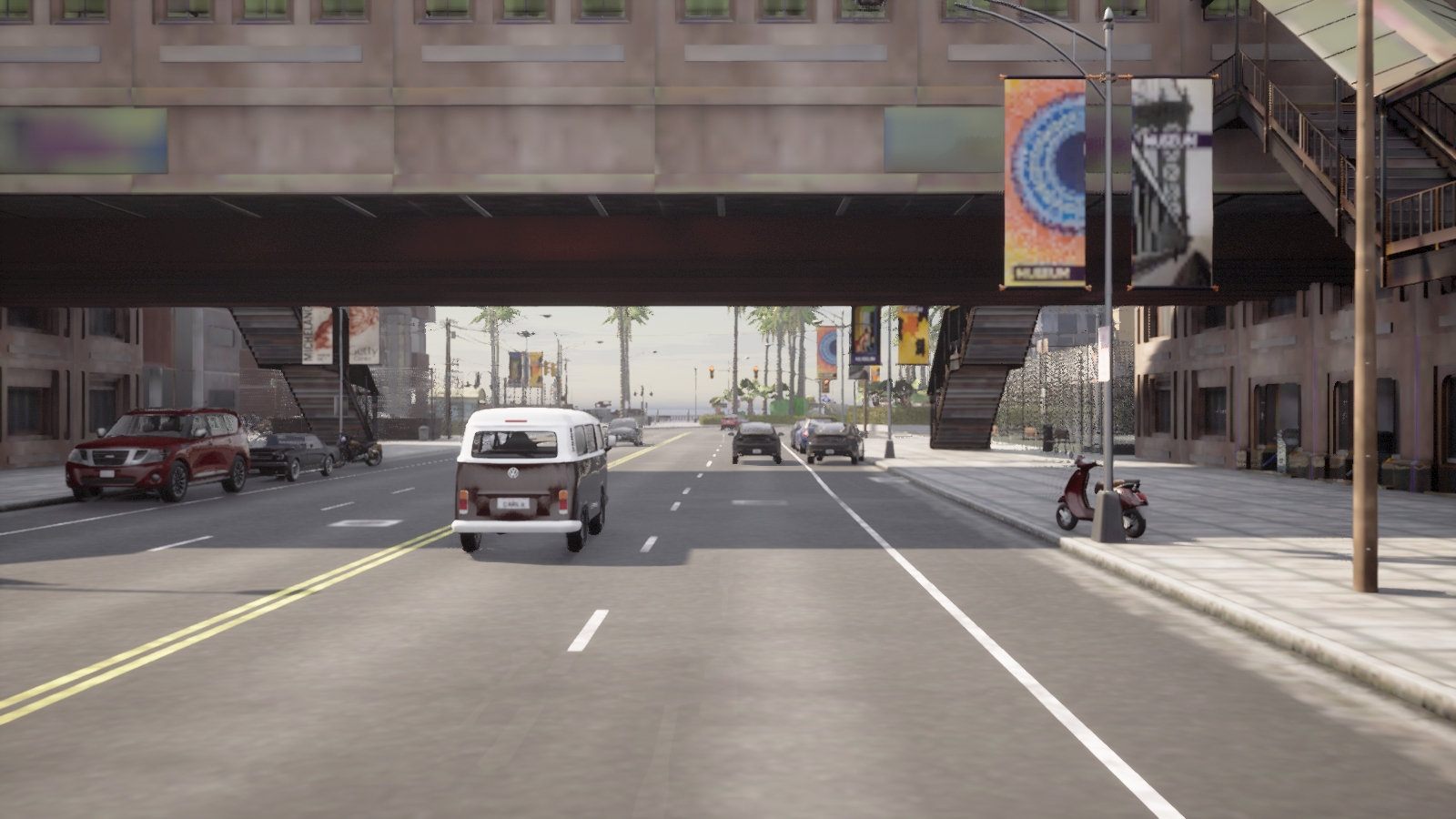}
        \caption{Right (10)}
    \end{subfigure}
    \begin{subfigure}[b]{0.15\textwidth}
        \centering
        \includegraphics[width=\textwidth]{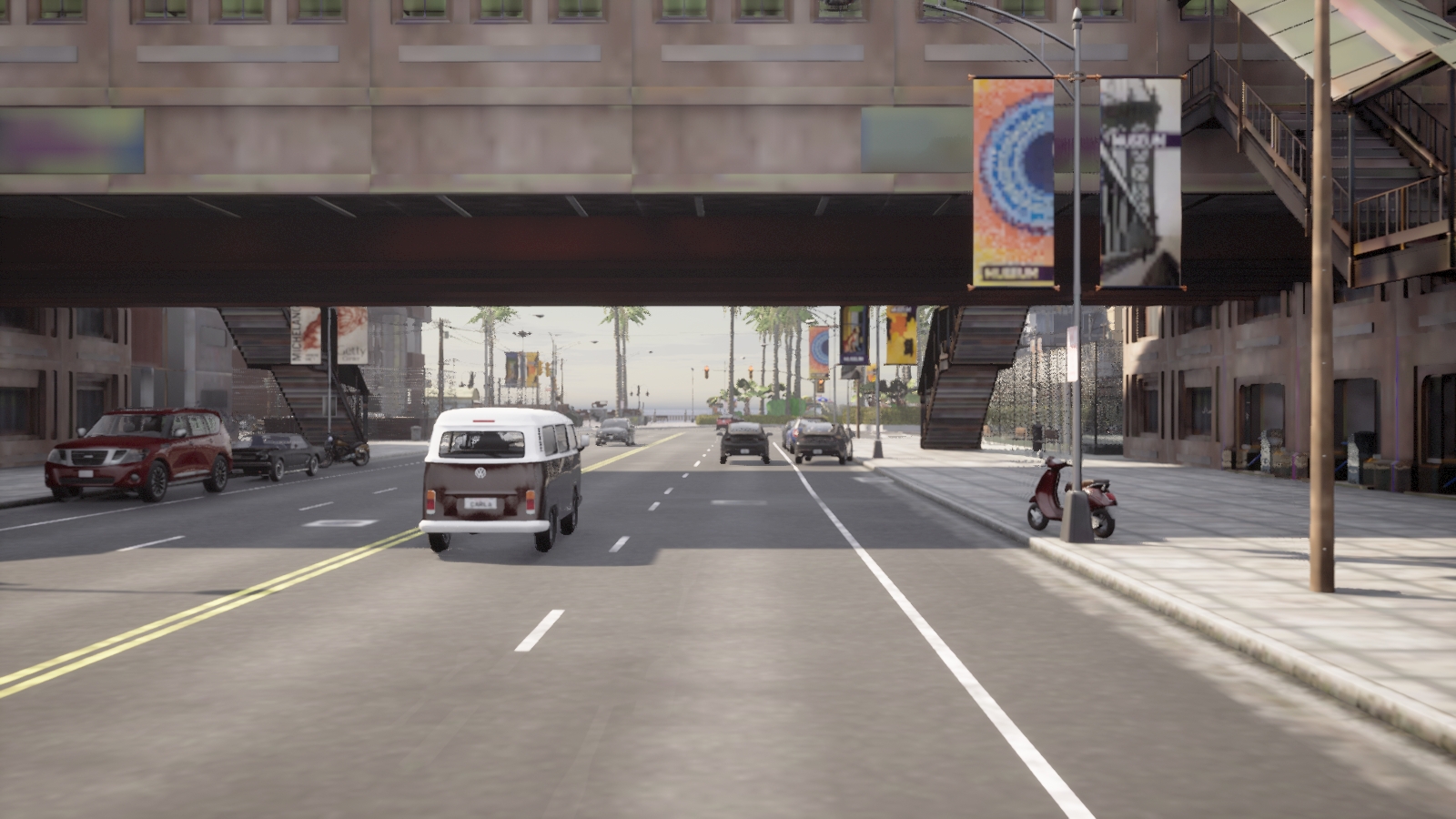}
        \caption{Right (54)}
    \end{subfigure}
    \begin{subfigure}[b]{0.15\textwidth}
        \centering
        \includegraphics[width=\textwidth]{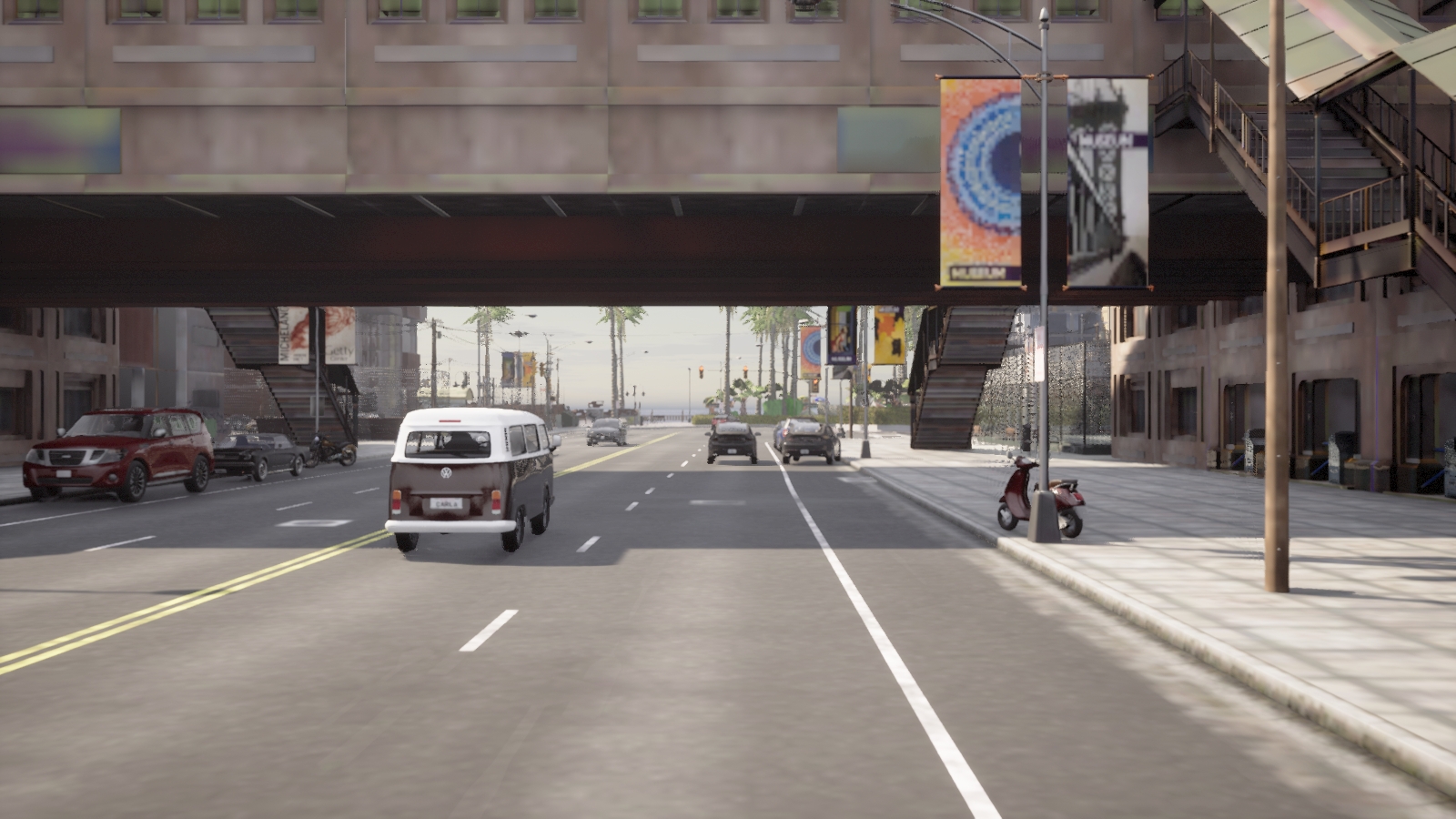}
        \caption{Right (100)}
    \end{subfigure}
    \begin{subfigure}[b]{0.15\textwidth}
        \centering
        \includegraphics[width=\textwidth]{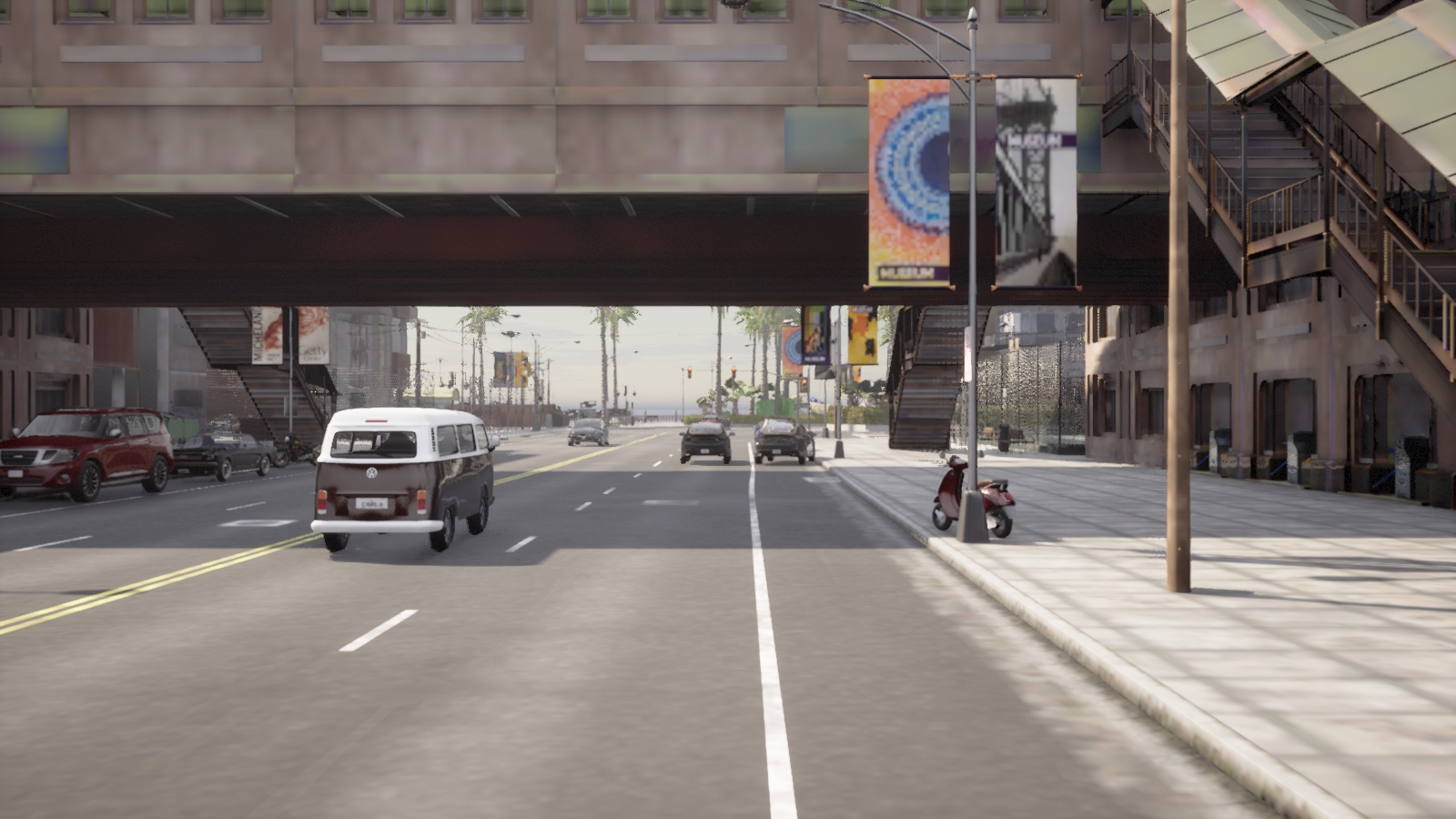}
        \caption{Right (200)}
    \end{subfigure}
    \begin{subfigure}[b]{0.15\textwidth}
        \centering
        \includegraphics[width=\textwidth]{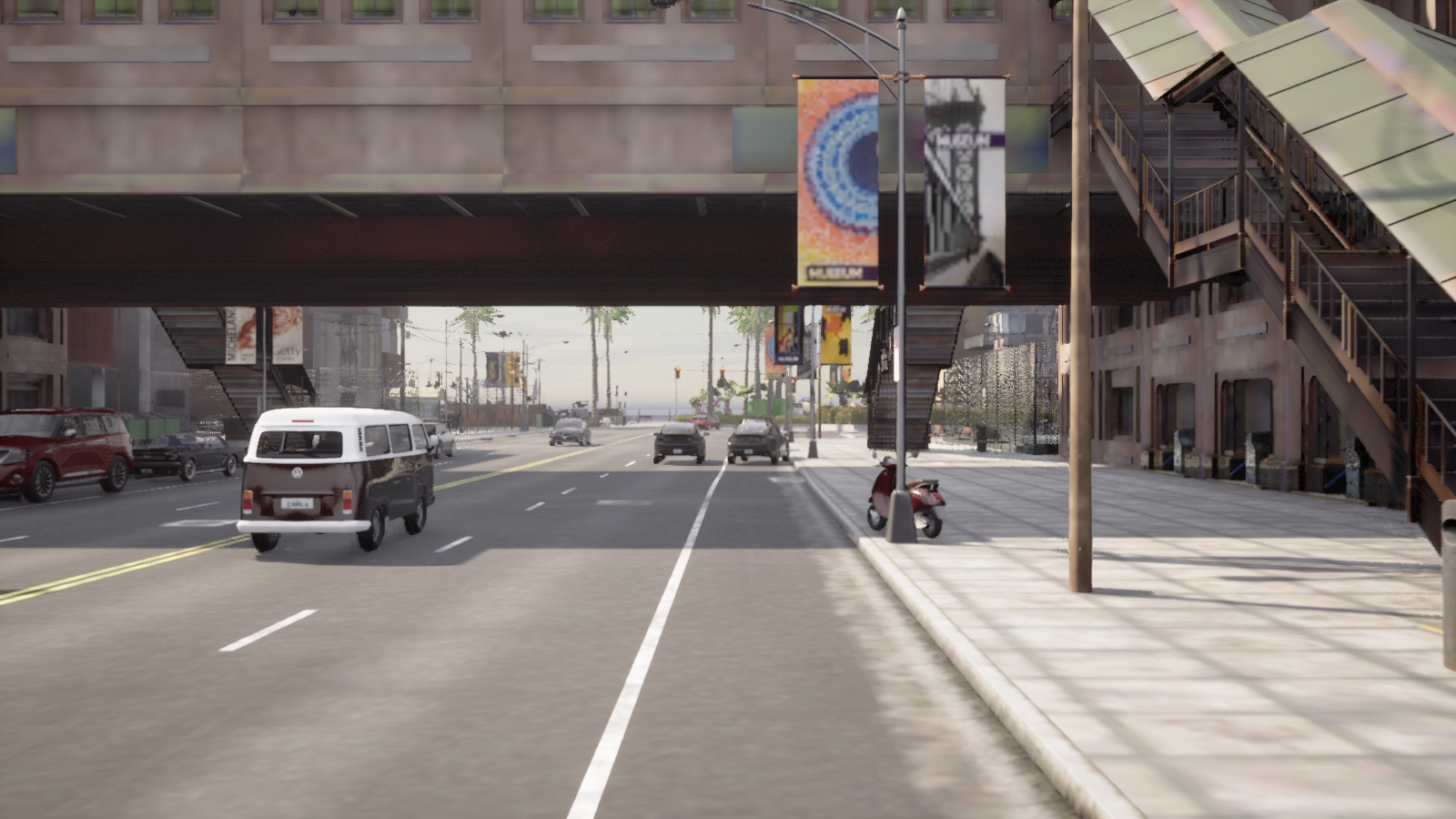}
        \caption{Right (300)}
    \end{subfigure}

    % 第二行图片
    \begin{subfigure}[b]{0.15\textwidth}
        \centering
        \includegraphics[width=\textwidth]{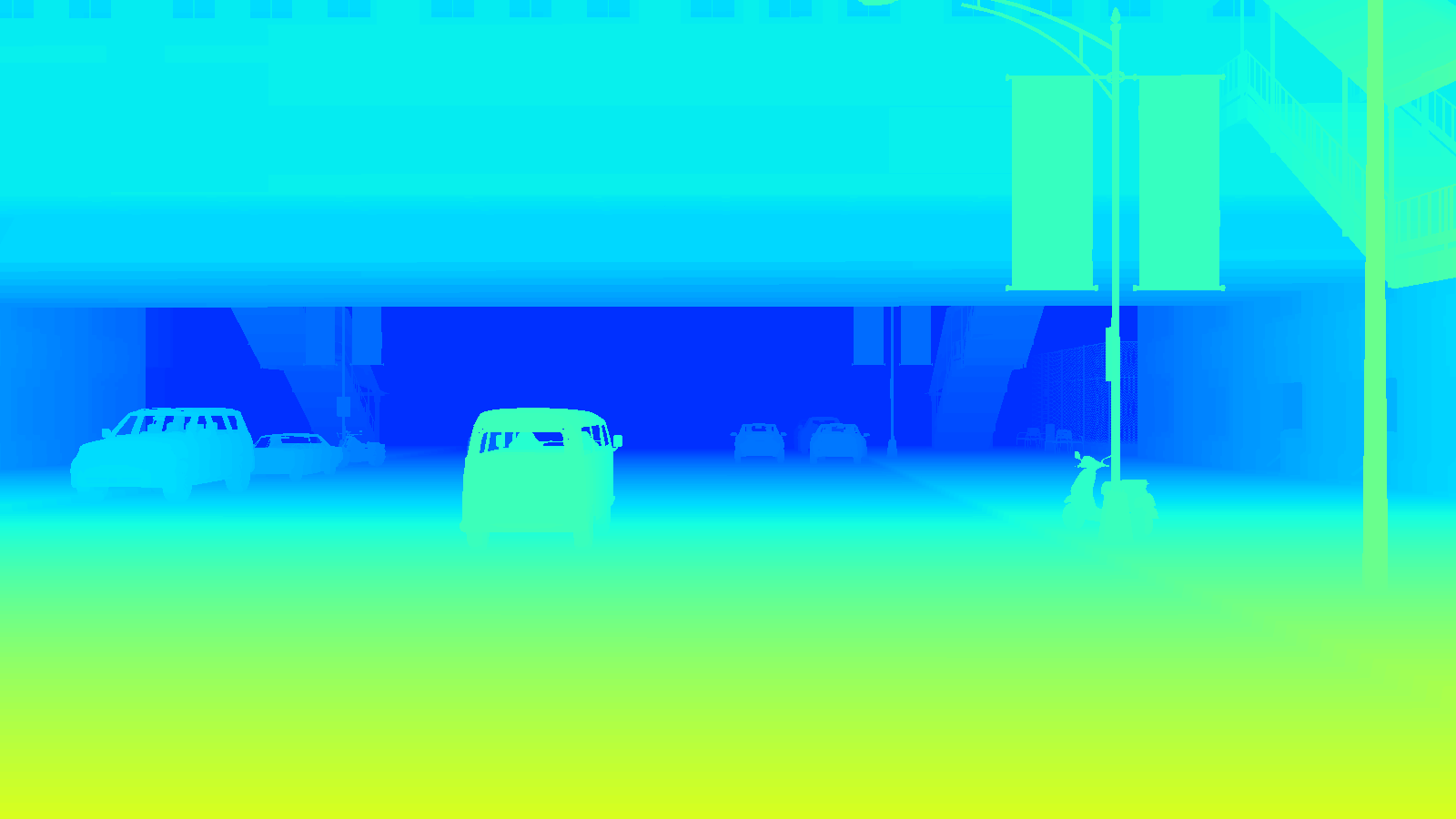}
        \caption{Depth}
    \end{subfigure}
    \begin{subfigure}[b]{0.15\textwidth}
        \centering
        \includegraphics[width=\textwidth]{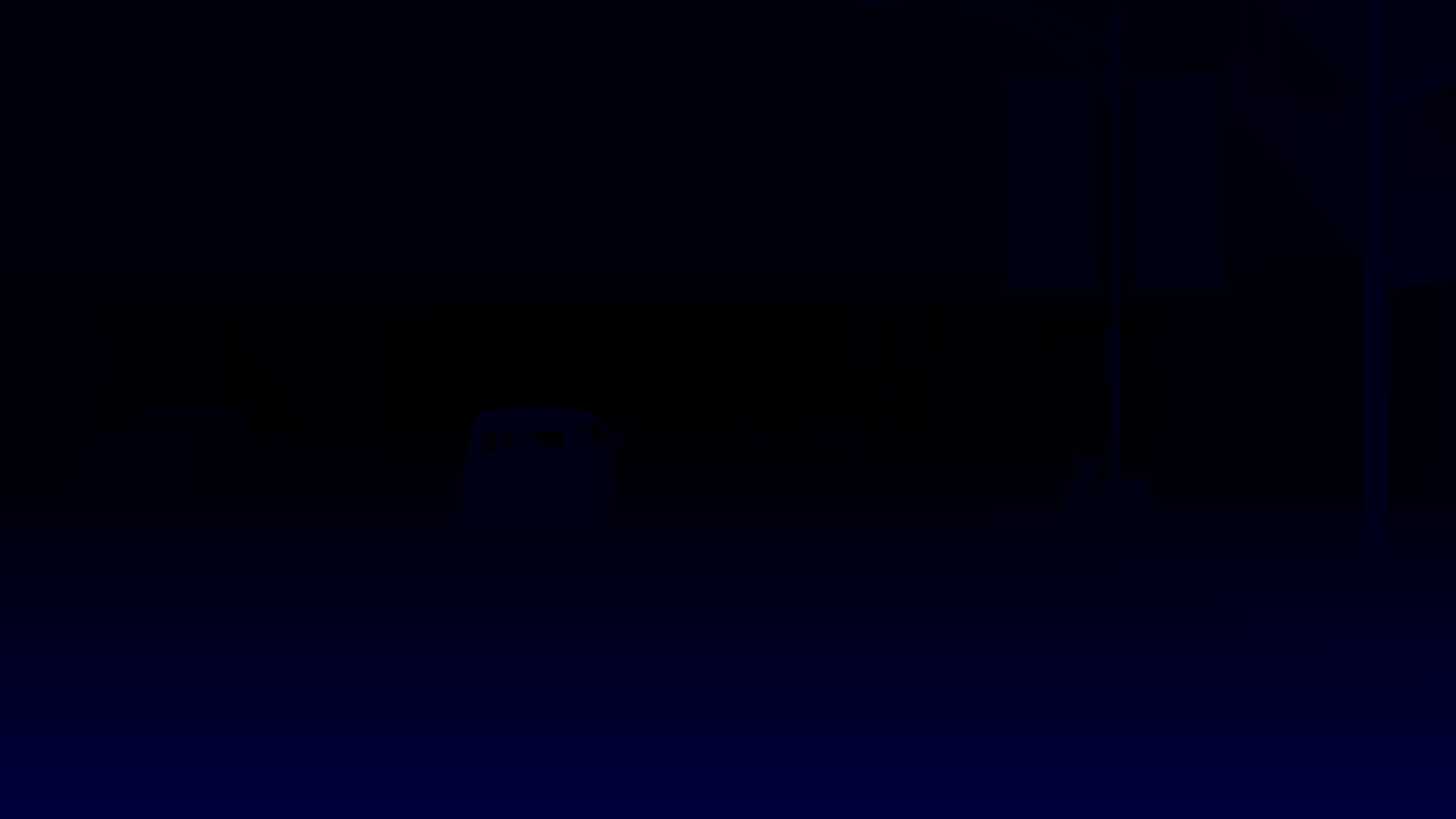}
        \caption{Disp (10)}
    \end{subfigure}
    \begin{subfigure}[b]{0.15\textwidth}
        \centering
        \includegraphics[width=\textwidth]{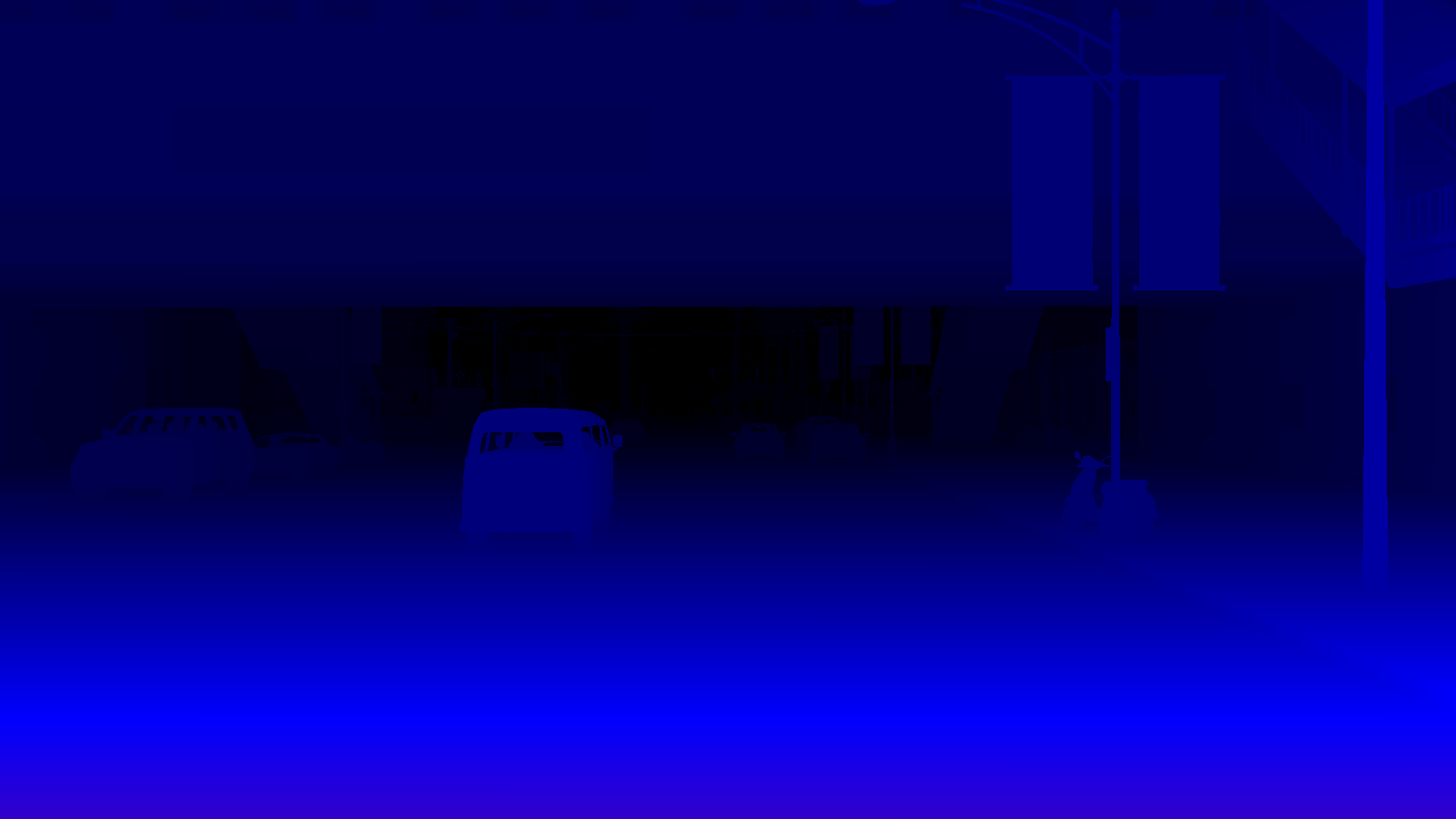}
        \caption{Disp (54)}
    \end{subfigure}
    \begin{subfigure}[b]{0.15\textwidth}
        \centering
        \includegraphics[width=\textwidth]{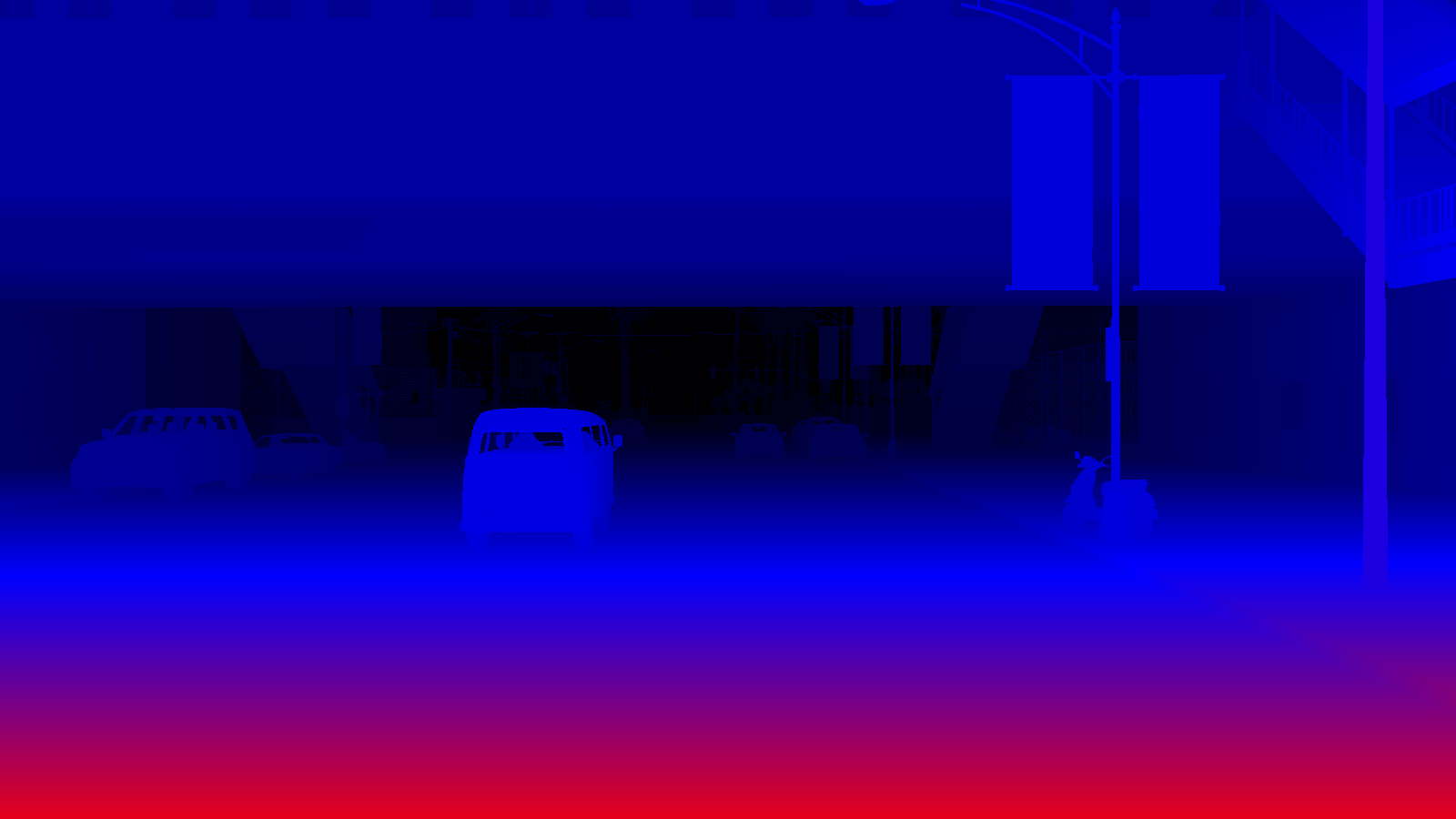}
        \caption{Disp (100)}
    \end{subfigure}
    \begin{subfigure}[b]{0.15\textwidth}
        \centering
        \includegraphics[width=\textwidth]{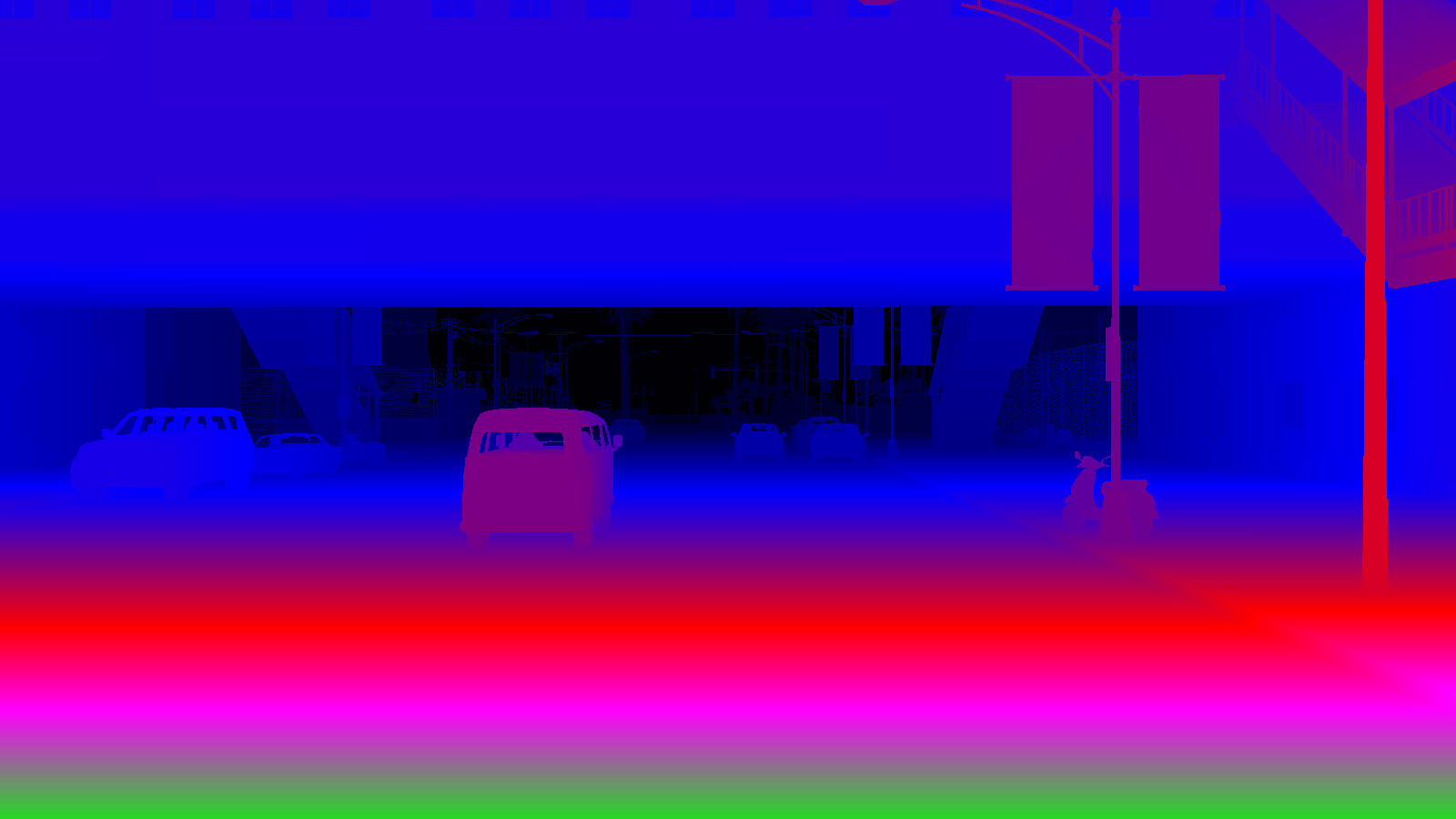}
        \caption{Disp (200)}
    \end{subfigure}
    \begin{subfigure}[b]{0.15\textwidth}
        \centering
        \includegraphics[width=\textwidth]{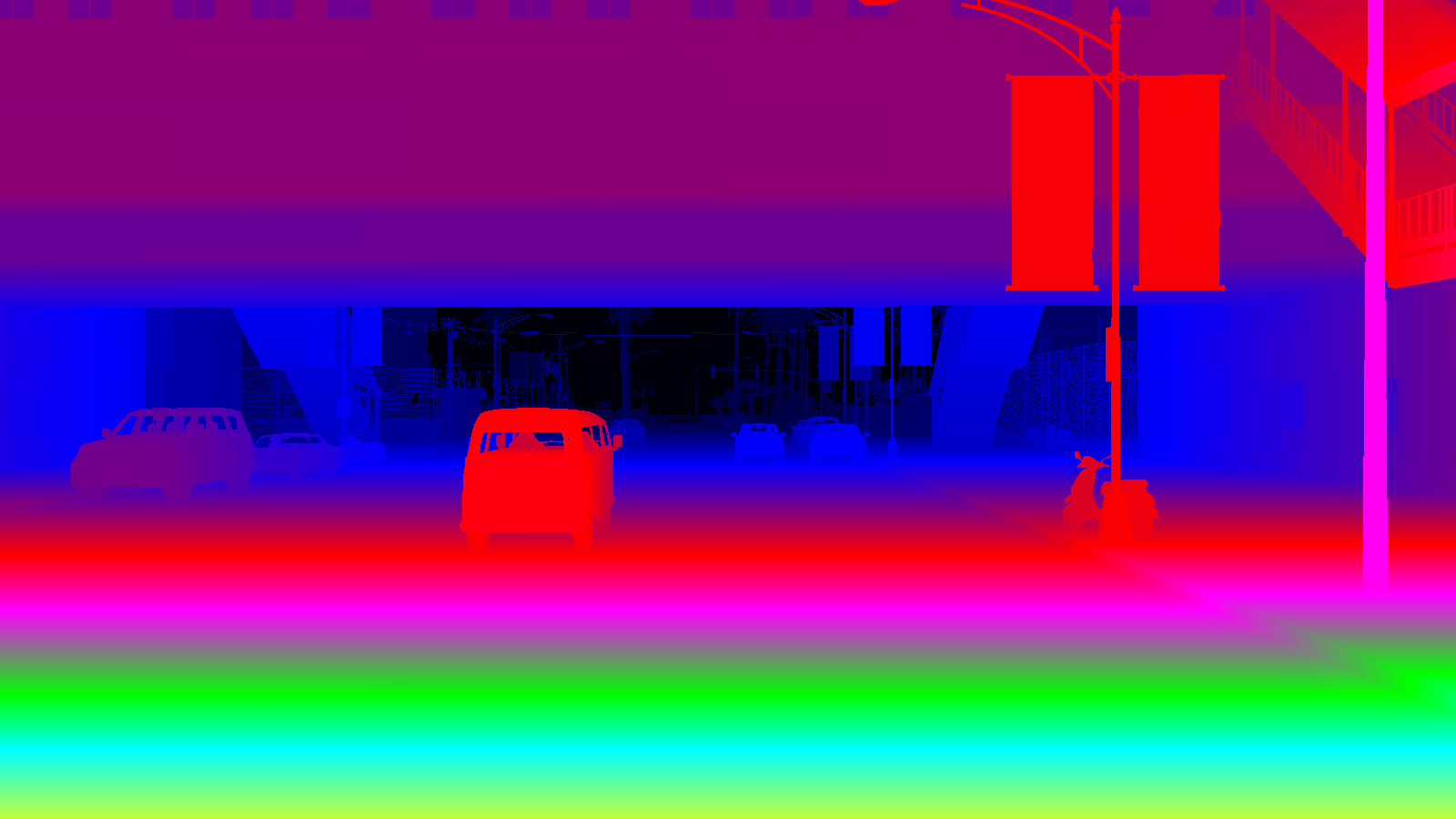}
        \caption{Disp (300)}
    \end{subfigure}

    % 第三行图片
    \begin{subfigure}[b]{0.15\textwidth}
        \centering
        \includegraphics[width=\textwidth]{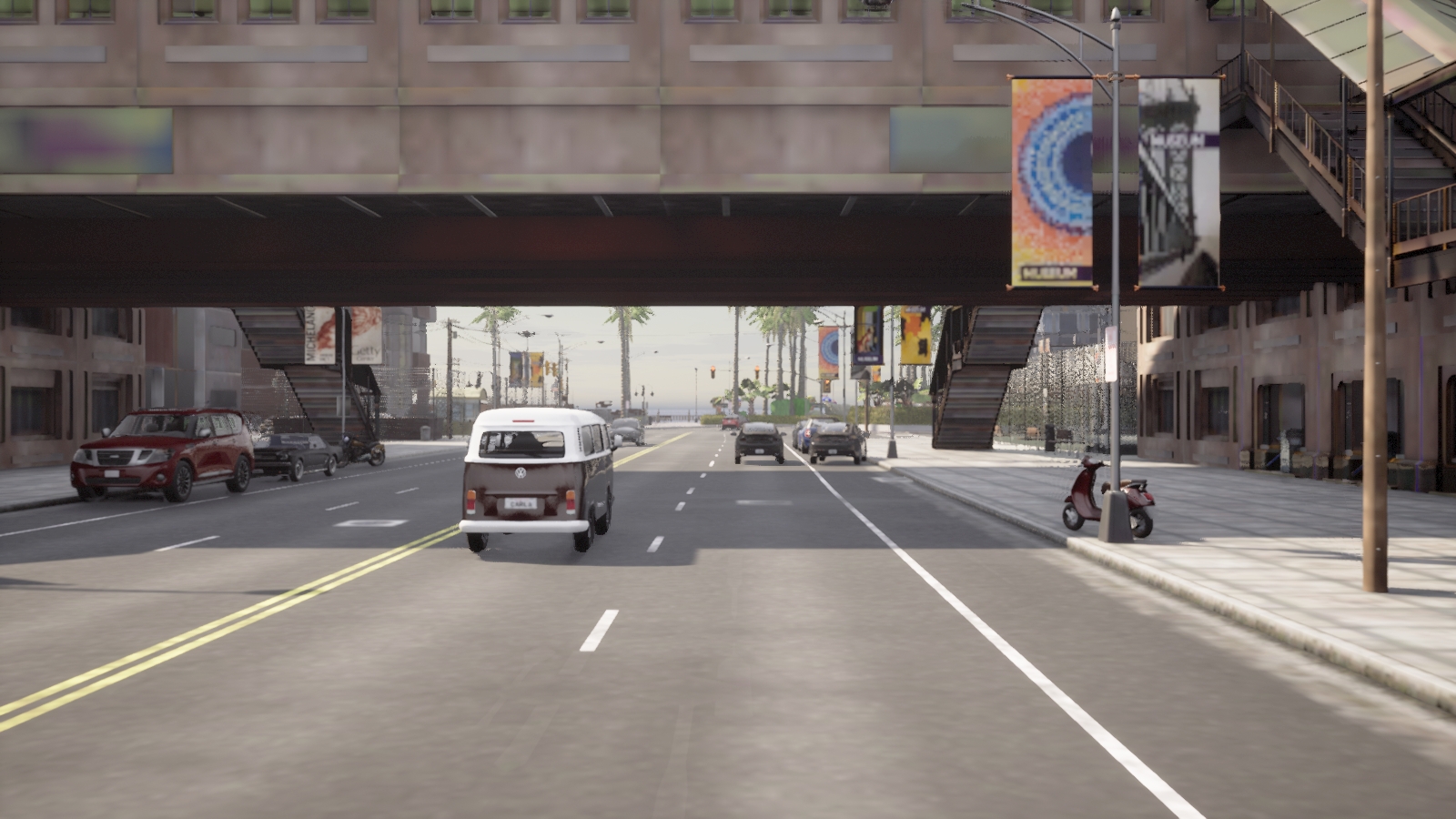}
        \caption{Normal}
    \end{subfigure}
    \begin{subfigure}[b]{0.15\textwidth}
        \centering
        \includegraphics[width=\textwidth]{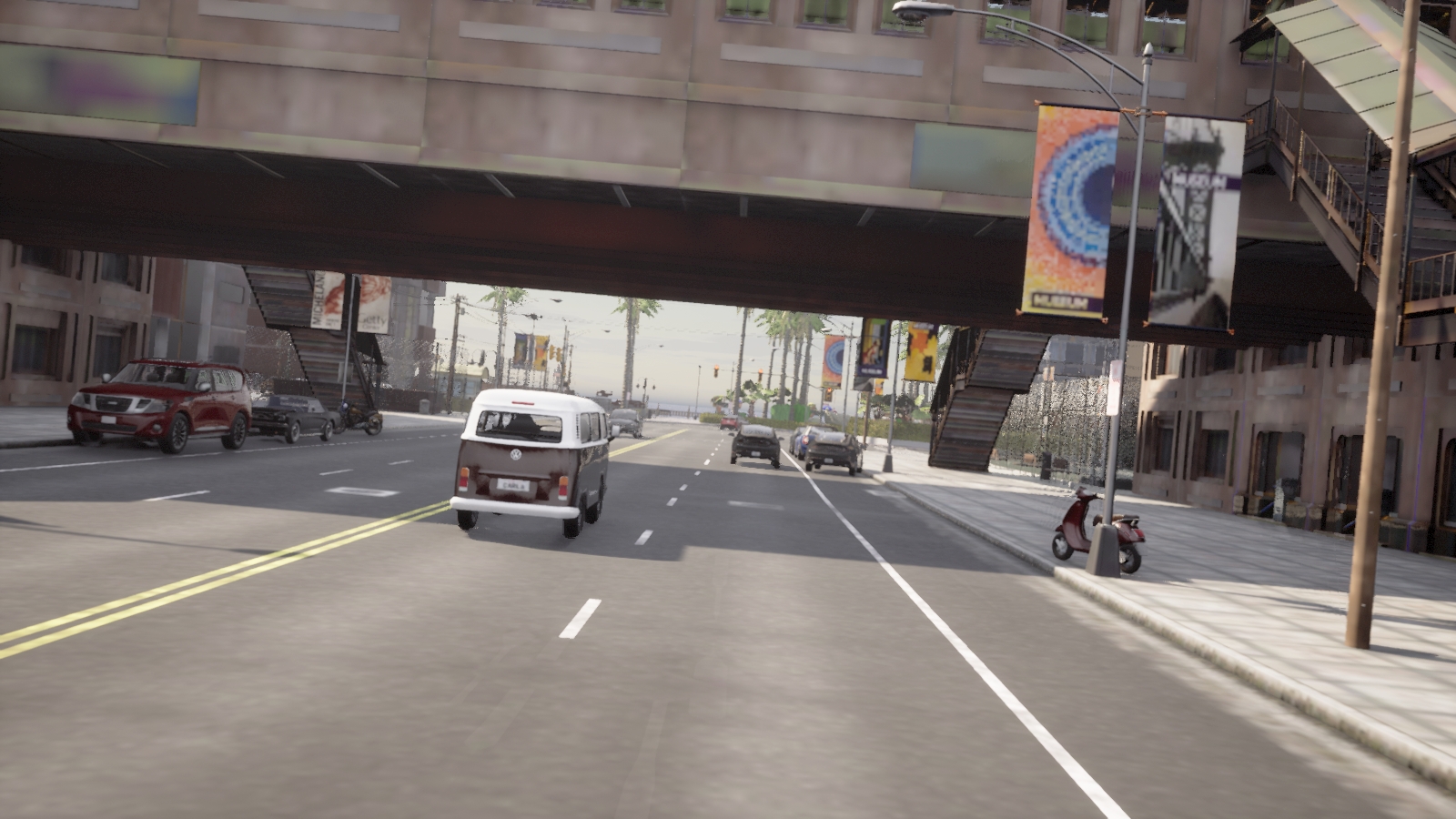}
        \caption{Roll $5^\circ$}
    \end{subfigure}
    \begin{subfigure}[b]{0.15\textwidth}
        \centering
        \includegraphics[width=\textwidth]{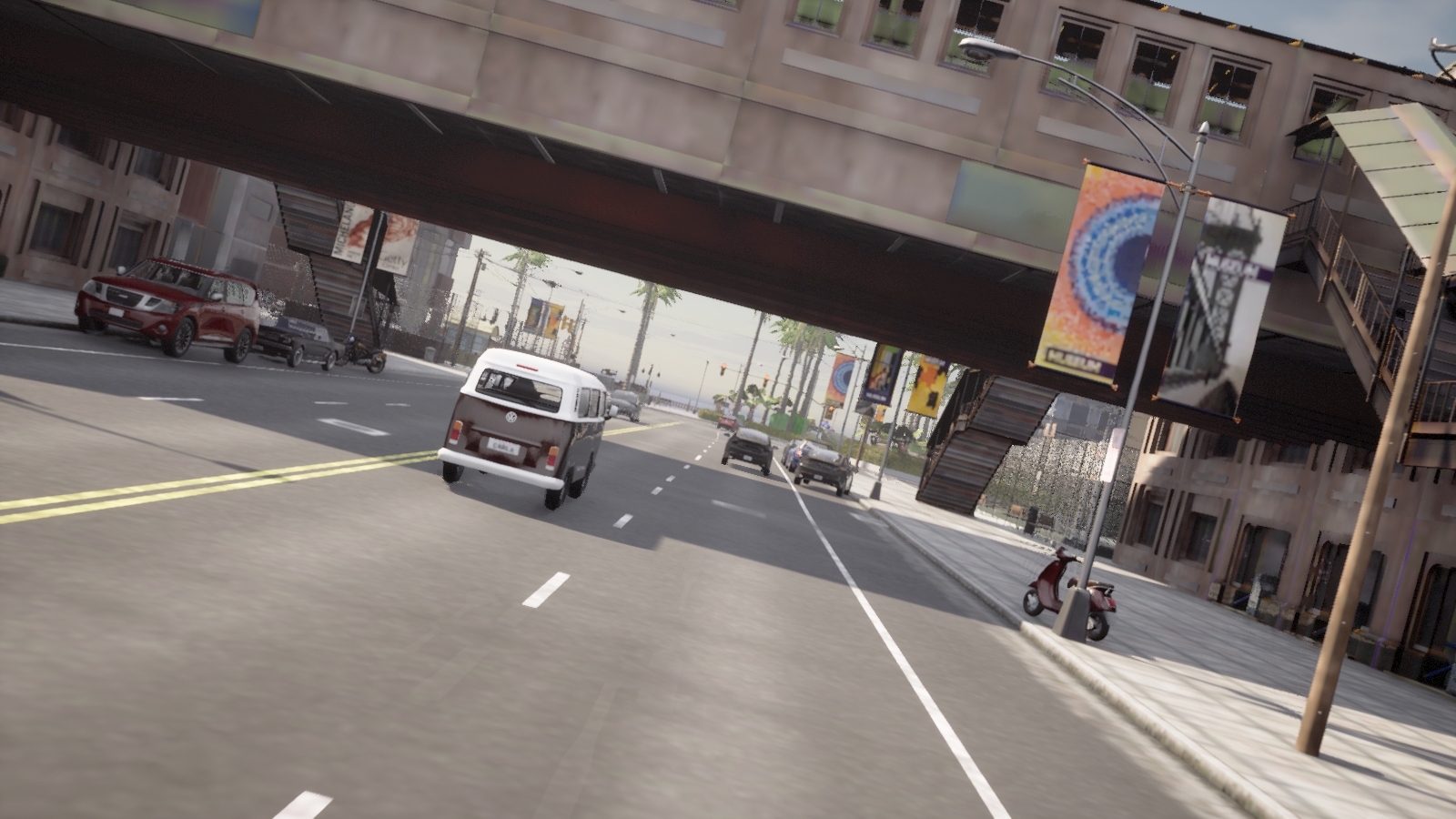}
        \caption{Roll $15^\circ$}
    \end{subfigure}
    \begin{subfigure}[b]{0.15\textwidth}
        \centering
        \includegraphics[width=\textwidth]{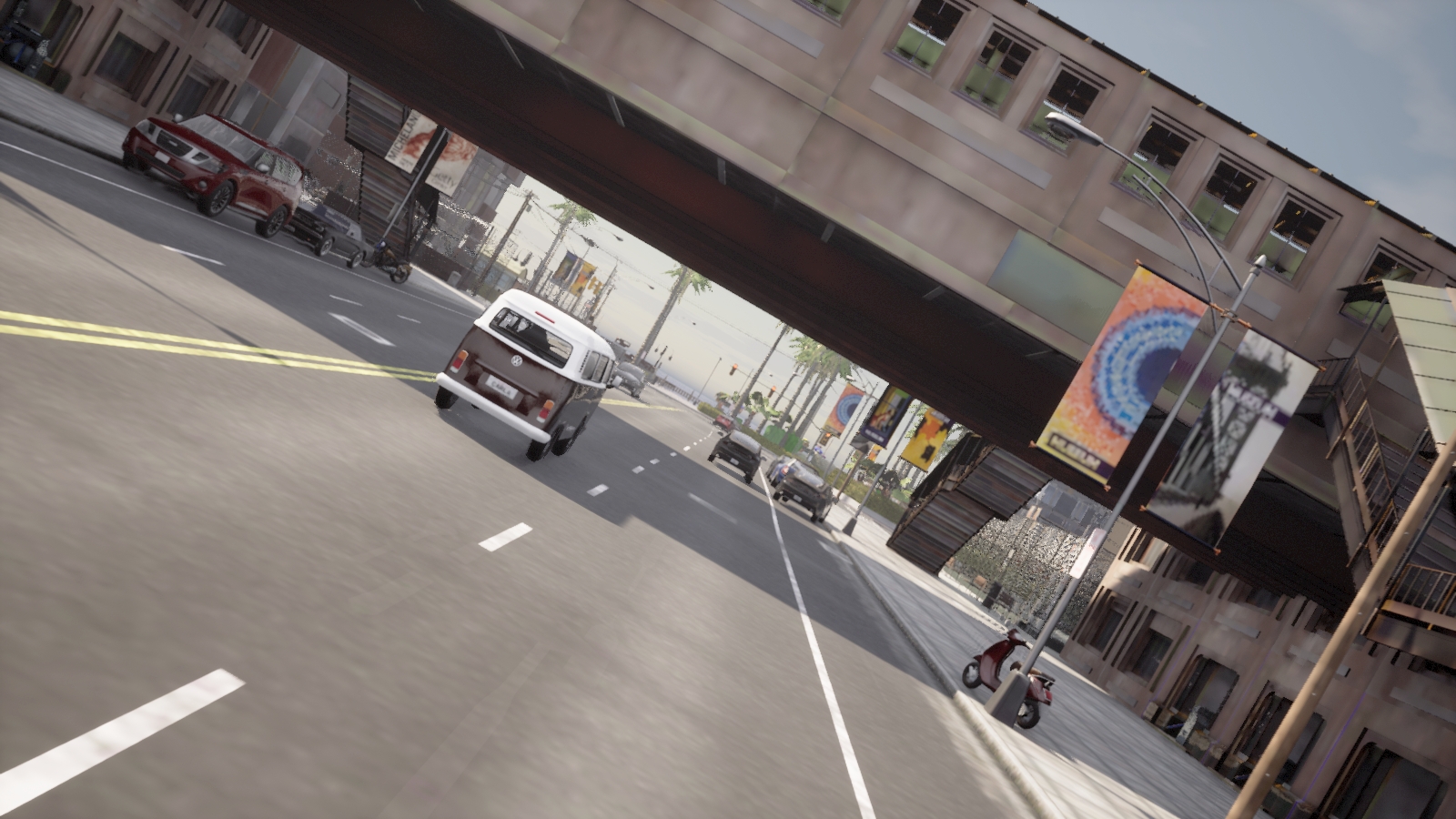}
        \caption{Roll $30^\circ$}
    \end{subfigure}
    \begin{subfigure}[b]{0.15\textwidth}
        \centering
        \includegraphics[width=\textwidth]{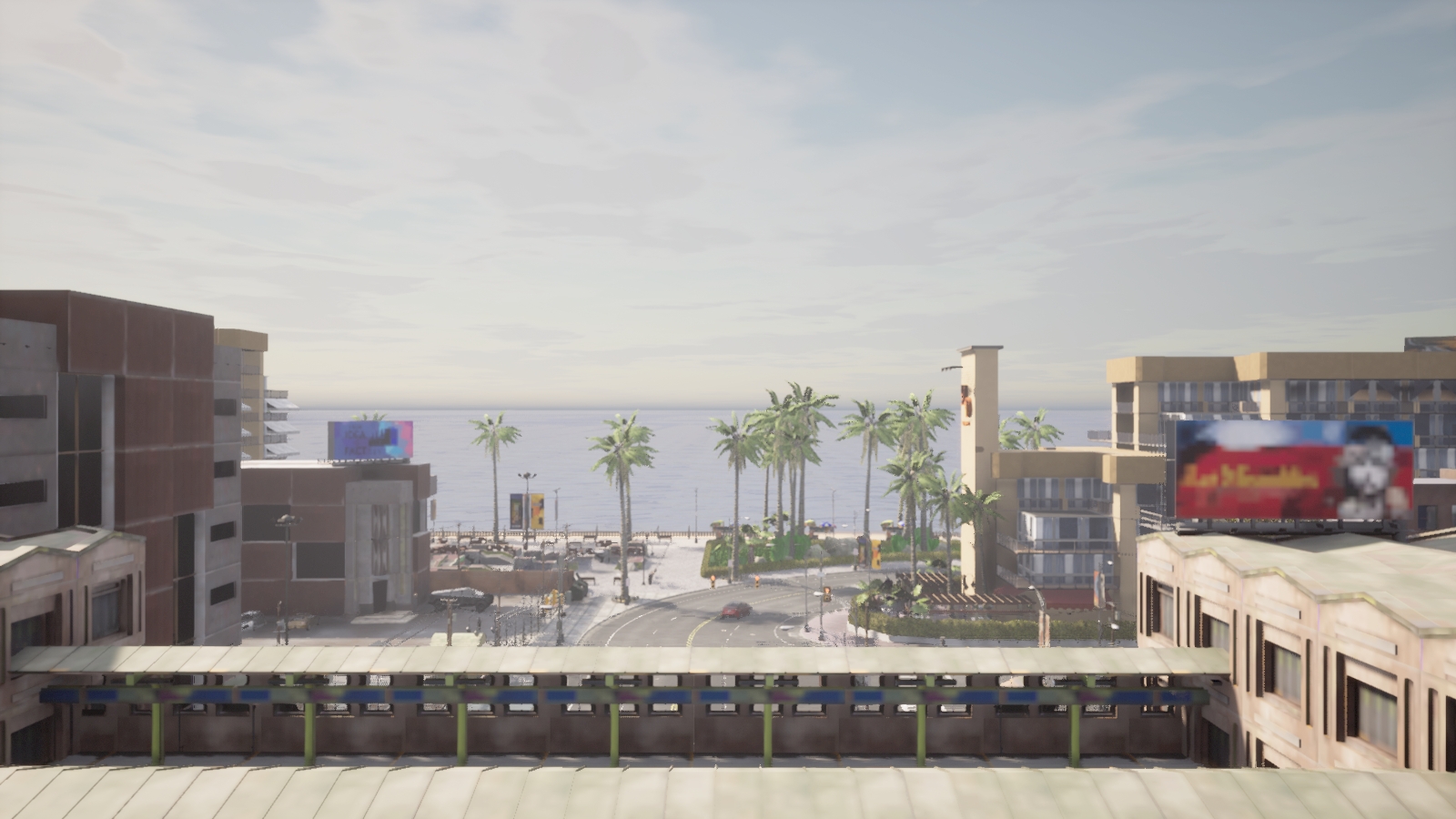}
        \caption{Height $20m$}
    \end{subfigure}
    \begin{subfigure}[b]{0.15\textwidth}
        \centering
        \includegraphics[width=\textwidth]{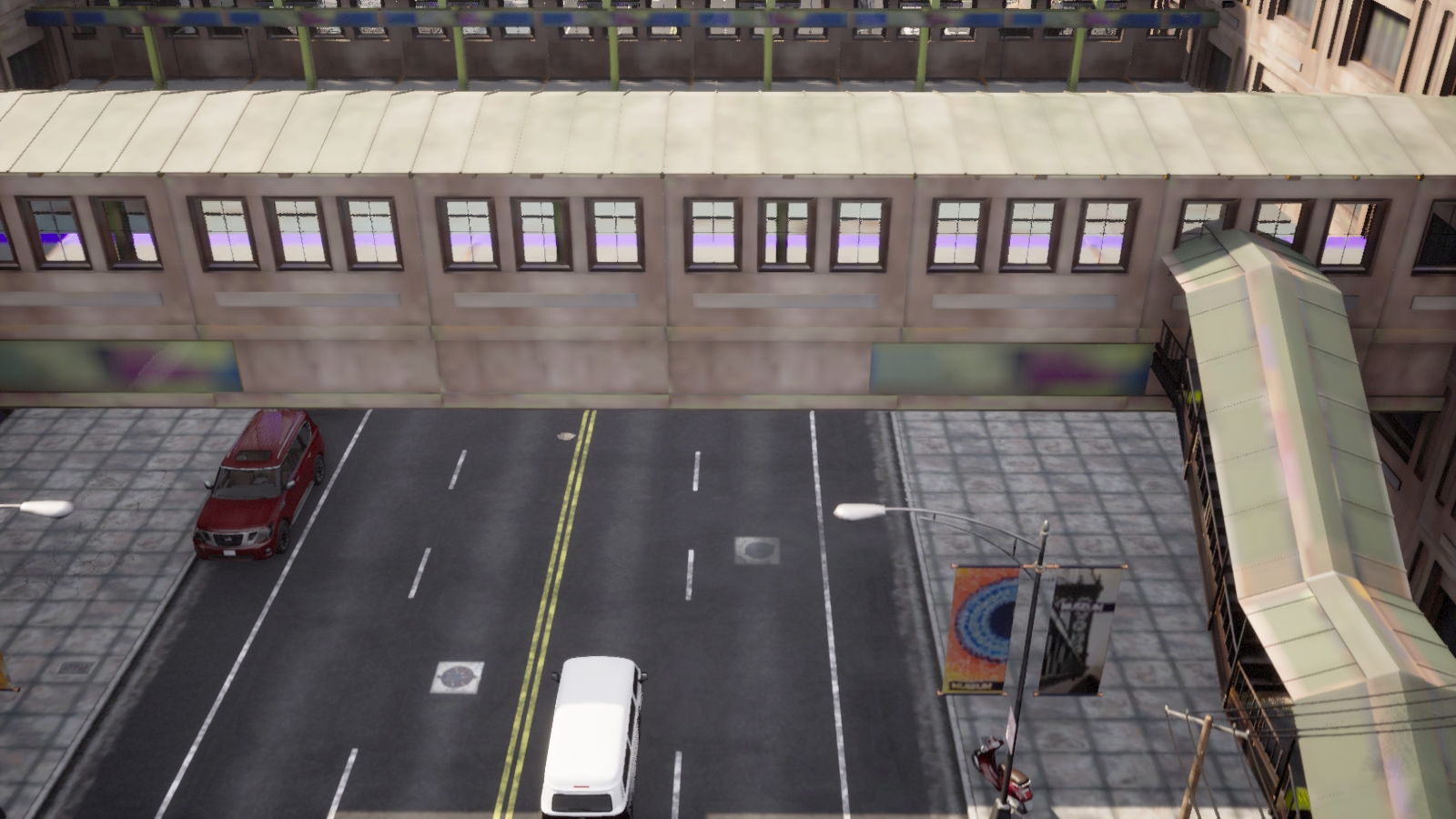}
        \caption{Pitch $-30^\circ$}
    \end{subfigure}
    
    \caption{\textbf{Overall of StereoCarla datasets.} The first row illustrates the left-eye image (1st column) and the right-eye image at varying baselines (2nd - 6th columns). The second row showcases the depth map (1st column) and corresponding Disp maps (2nd - 6th columns). The third rows depict left images from varied horizontal viewing angles and elevated viewpoints.} 
    \label{stereocarla}
    % \vspace{-5mm}
\end{figure*}

\subsection{Stereo Matching}

With the rapid advancement of deep learning, learning-based stereo matching methods have largely replaced traditional optimization-based approaches. GCNet~\cite{gcnet} constructs a 4D cost volume and aggregates costs via 3D CNN neural networks, laying the foundation for modern deep stereo networks. Building upon this idea, many models such as PSMNet~\cite{psmnet2018}, GwcNet~\cite{gwcnet2019}, ACVNet~\cite{acvnet}, and GA-Net~\cite{ganet2019} have progressively improved accuracy by enhancing feature extraction and cost aggregation modules.
Inspired by the success of RAFT~\cite{teed2020raft} in optical flow estimation, a series of iterative optimization-based stereo methods have emerged, including RAFT-Stereo~\cite{raftstereo}, IGEV~\cite{xu2023iterative}, StereoBase~\cite{guo2023openstereo}, and IGEV++~\cite{xu2024igev++}. These models refine disparity maps through multiple update steps, achieving state-of-the-art accuracy on several benchmarks. However, the computational complexity and inference latency of these methods remain major bottlenecks, limiting their deployment in real-time applications.
To address this trade-off, recent efforts~\cite{xu2020aanet,wang2020fadnet,shamsafar2022mobilestereonet,guo2025lightstereo} have explored lightweight architectures that construct 3D cost volumes with 2D CNN refinement, aiming to balance performance and efficiency. 
In addition, there has been increasing interest in video stereo matching~\cite{jing2024matchstereovideos, jing2024matchsv}, which leverages temporal consistency across frames to improve depth estimation. Jing et al. \cite{jing2024matchstereovideos} introduce a bidirectional alignment operation that enforces temporal coherence in disparity predictions, while \cite{jing2024matchsv} leverages monocular video depth priors to enhance the robustness of features.

\section{StereoCarla Datasets}

To expand the diversity and quantity of existing stereo-matching datasets, we utilized the CARLA~\cite{carla} simulator to collect new synthetic stereo data. 
Compared to previous stereo datasets, StereoCarla offers more varied settings, providing different baselines and novel camera configurations that enhance the richness of stereo data. 
Table~\ref{tab:datasets_summary} provides a comparative summary of existing labeled stereo datasets and our proposed StereoCarla dataset.
Fig.~\ref{stereocarla} illustrates some examples of our collected dataset, showcasing the diverse baselines, camera angles, and scene variations. Below, we detail the major design considerations:

\begin{figure}
    \centering
    \includegraphics[width=0.9\linewidth]{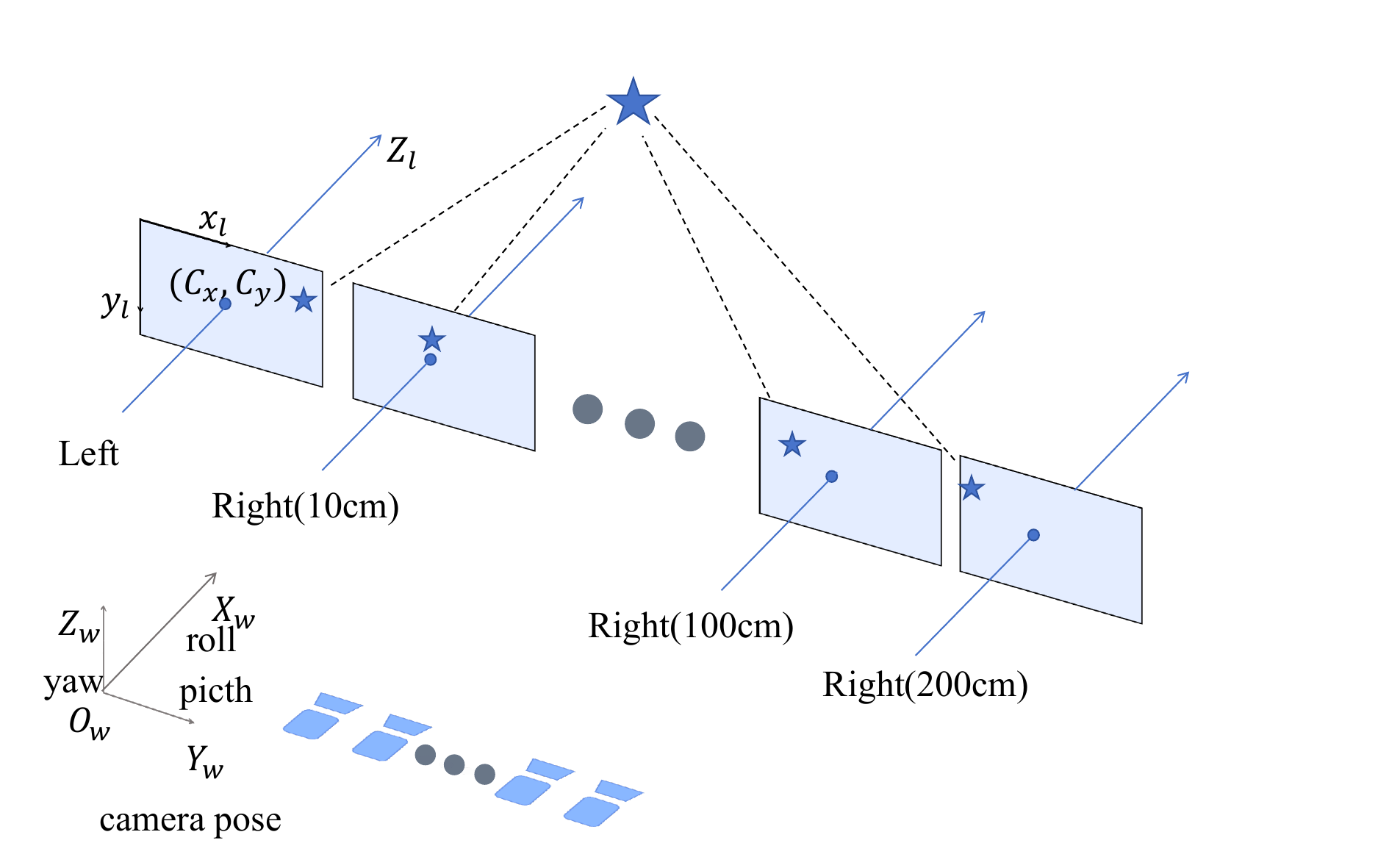}
    \caption{\textbf{Camera Settings for data generation.}}
    \label{fig:camera}
    \vspace{-5mm}
\end{figure}

\textbf{Multiple Baselines.} As shown in Fig.~\ref{fig:camera}, we collected data with baseline distances set at 10 cm, 54 cm, 100 cm, 200 cm, and 300 cm, which are much broader in range compared to existing datasets. These baseline variations allow the model to better generalize to scenarios where the distance between the two cameras can differ greatly, such as switching between different hardware configurations in real-world applications. 

\textbf{Horizontal Viewing Angles.} We collected data at four different horizontal viewing angles. Specifically, we captured images at a direct horizontal view (0 degrees) and slight and moderate rotations of 5, 15, and 30 degrees. These different orientations provide more robust data, as they simulate variations that occur naturally when a stereo camera system is in motion or when changes in scene viewpoint occur.

\textbf{Elevated Viewpoints.} Furthermore, to simulate a diverse set of environments and perspectives, we positioned the stereo camera pair at an elevated height of 10 meters above the vehicle, capturing the scenes from both a horizontal view and at a 30-degree downward tilt. This unique perspective provides more information about the overall layout of the scene, which can be particularly useful in understanding both near-ground details and broader contextual elements that are missing in most existing datasets.

\textbf{High Resolution.} Each scene in our dataset consists of left and right images, with a resolution of $1600 \times 900$ pixels, along with corresponding dense disparity maps, ensuring comprehensive ground-truth information for both training and evaluating state-of-the-art stereo models. 

\textbf{Weather Variations.} We captured stereo data under clear, cloudy, foggy, humid, night, storm, and sunset conditions. These variations introduce realistic challenges such as reduced visibility, altered lighting, and reflections, which are crucial for training robust stereo-matching models.

By incorporating the above elements, our dataset aims to address key limitations in existing stereo datasets, particularly the lack of varied viewpoints and baseline distances.

\begin{table}[t]
    \centering
    \setlength\tabcolsep{1.5pt}
    \renewcommand\arraystretch{1.1}
    \caption{\textbf{StereoCarla samples distribution.} Number of Samples Collected Across Different Towns and Camera Configurations. Each camera configuration includes baselines of 10, 54, 100, 200, and 300cm.}
    \scalebox{0.85}{
    \begin{tabular}{l|ccccccc|c|c}
        \toprule
        \rowcolor{gray!20}
        \textbf{Setting}& \textbf{town01} & \textbf{town02} & \textbf{town03} & \textbf{town04} & \textbf{town05} & \textbf{town06} & \textbf{town07} & \textbf{town10}&\textbf{Sum} \\ 
        \midrule
        \textbf{Normal} & 25,010 & 25,010 & 12,175 & 12,475 & 12,500 & 12,485 & 12,300 & 12,475 & 124,430 \\
        \rowcolor{gray!10}
\textbf{Roll05} & 12,505 & 12,505 & 12,505 & 12,495 & 25,005 & 12,500 & 12,465 & 12,490 & 112,470 \\
\textbf{Roll15} & 12,500 & 12,505 & 12,500 & 12,490 & 12,495 & 24,985 & 12,430 & 12,500 & 112,405 \\
\rowcolor{gray!10}
\textbf{Roll30} & 12,500 & 12,505 & 12,505 & 12,500 & 12,500 & 12,500 & 25,000 & 24,995 & 125,005 \\
\textbf{Pitch00} & 12,505 & 12,505 & 24,995 & 12,495 & 12,495 & 12,495 & 12,500 & 12,495 & 112,485 \\ 
\rowcolor{gray!10}
\textbf{Pitch30} & 6,795  & 12,505 & 12,500 & 24,990 & 12,495 & 12,500 & 12,420 & 12,435 & 106,640 \\
\midrule
\textbf{Sum}& 81,815 & 87,535 & 87,180 & 87,445 & 87,490 & 87,465 & 87,115 & 87,390 & 693,435 \\
        \bottomrule
    \end{tabular}
    }
    \vspace{-0.5cm}
    \label{tab:stereo_data}
\end{table}

\section{Experiments}

\subsection{Implementation Details}

We conduct experiments based on OpenStereo~\cite{guo2023openstereo}. 
NMRF-Stereo~\cite{NMRFStereo} with a SwinTransformer backbone is adopted as the baseline model due to its superior performance and inference speed. All experiments use the AdamW optimizer with a batch size of 16. During training, we apply data augmentation techniques—including random cropping ($352 \times 640$), color jitter, and random erasing—to improve model robustness. 
We train the model for 31,250 iterations when using single datasets, and extend training to 81,000 iterations for mixed-dataset scenarios, ensuring sufficient learning from diverse combined data. Regarding the learning rate schedule, we adopt OneCycleLR scheduling with a maximum learning rate of 0.001 for training on SceneFlow~\cite{sceneflow}, while for fine-tuning on other datasets, it is reduced to 0.0005.

\begin{table}[t]
 % MG refers to Marigold~\cite{Marigold}.
  \small
  \centering
  \caption{\textbf{\textbf{Zero-shot performance} when fine-tuning on different training sets.} Relative to baseline (top row), \textcolor{OliveGreen}{green}/\textcolor{red}{red} shows performance improvement/decline. Best results in \textbf{bold}, second-best \underline{underlined}.}
  \setlength\tabcolsep{5pt}
  \renewcommand\arraystretch{1.1}
  % \ra{1.05}
\scalebox{0.8}{  
\begin{tabular}{l||cccc|cc}
\thickhline
\rowcolor{gray!20}
\textbf{Dataset} & \textbf{K12} & \textbf{K15} & \textbf{Midd} & \textbf{E3D} & \textbf{Mean} & \textbf{Rank} \\
\hline\hline
SF~\cite{sceneflow} & 8.67 & 7.46 & 16.36 & 23.46 & 13.99 & - \\
\hline

\rowcolor{cyan!10}
\textbf{SF$\rightarrow$StereoCarla (Ours)} & \underline{\textcolor{OliveGreen}{4.11}} & \textcolor{OliveGreen}{4.87} & \textbf{\textcolor{OliveGreen}{9.12}} & \textbf{\textcolor{OliveGreen}{3.17}} & \textbf{5.32} & 1 \\

SF$\rightarrow$Tartanair~\cite{wang2020tartanair} & \textcolor{OliveGreen}{4.16} & \textcolor{OliveGreen}{4.71} & \textcolor{OliveGreen}{13.95} & \underline{\textcolor{OliveGreen}{5.25}} & \underline{7.02} & 2 \\

\rowcolor{gray!10} 
SF$\rightarrow$CREStereo~\cite{Crestereo} & \textcolor{OliveGreen}{8.01} & \textcolor{OliveGreen}{6.18} & \textcolor{OliveGreen}{13.73} & \textcolor{OliveGreen}{5.75} & 8.42 & 3 \\

SF$\rightarrow$Spring~\cite{mehl2023spring} & \textcolor{OliveGreen}{6.59} & \textcolor{OliveGreen}{6.23} & \textcolor{OliveGreen}{16.04} & \textcolor{OliveGreen}{6.96} & 8.96 & 4 \\

\rowcolor{gray!10}
SF$\rightarrow$Sintel~\cite{Sintel} & \textcolor{OliveGreen}{6.09} & \textcolor{OliveGreen}{6.28} & \textcolor{red}{19.28} & \textcolor{OliveGreen}{6.18} & 9.46 & 5 \\

SF$\rightarrow$DynamicReplica~\cite{karaev2023dynamicstereo} & \textcolor{red}{11.84} & \textcolor{red}{15.36} & \textcolor{OliveGreen}{12.84} & \textcolor{OliveGreen}{5.32} & 11.34 & 6 \\

\rowcolor{gray!10}
SF$\rightarrow$FallingThings~\cite{tremblay2018falling} & \textcolor{OliveGreen}{4.28} & \underline{\textcolor{OliveGreen}{4.23}} & \textcolor{OliveGreen}{13.17} & \textcolor{red}{27.93} & 12.40 & 7 \\

SF$\rightarrow$Instereo2K~\cite{bao2020instereo2k} & \textcolor{red}{13.33} & \textcolor{red}{15.21} & \underline{\textcolor{OliveGreen}{11.75}} & \textcolor{OliveGreen}{11.23} & 12.88 & 8 \\

\rowcolor{gray!10}
SF$\rightarrow$VirtualKitti2~\cite{cabon2020virtual} & \textbf{\textcolor{OliveGreen}{3.96}} & \textbf{\textcolor{OliveGreen}{4.00}} & \textcolor{red}{22.23} & \textcolor{red}{73.78} & 25.99 & 9 \\

SF$\rightarrow$UnrealStereo4K~\cite{tosi2021smd} & \textcolor{red}{8.68} & \textcolor{OliveGreen}{6.90} & \textcolor{red}{44.98} & \textcolor{red}{64.51} & 31.27 & 10 \\

\thickhline
\end{tabular}}

% \caption*{Note: The total training iter is 31,250 for the fine-tuned models in the table. NMRF-Stereo-SwinT. }
\label{tab:onetraincolor}
\vspace{-3mm}
\end{table}

\begin{figure*}[htbp]
    \centering
    % 第一行图片
    \begin{adjustbox}{rotate=90}\hspace{0.1mm}K12\end{adjustbox}
    \begin{subfigure}[b]{0.13\textwidth}
        \centering
        \includegraphics[width=\textwidth]{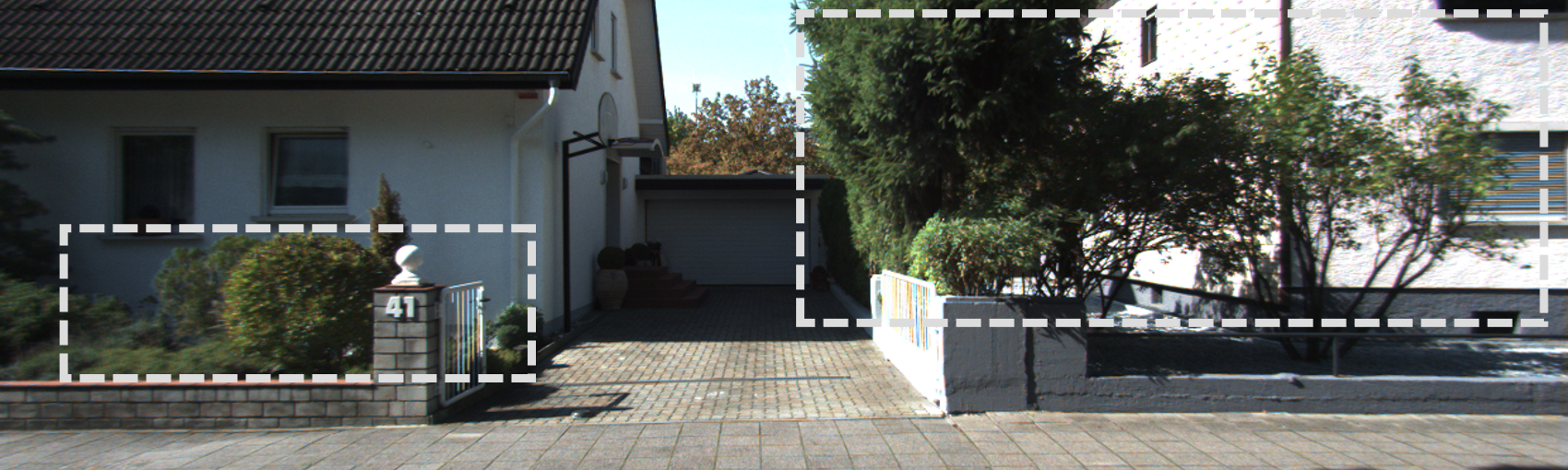}
        % \caption{StereoCarla}
    \end{subfigure}
    \begin{subfigure}[b]{0.13\textwidth}
        \centering
        \includegraphics[width=\textwidth]{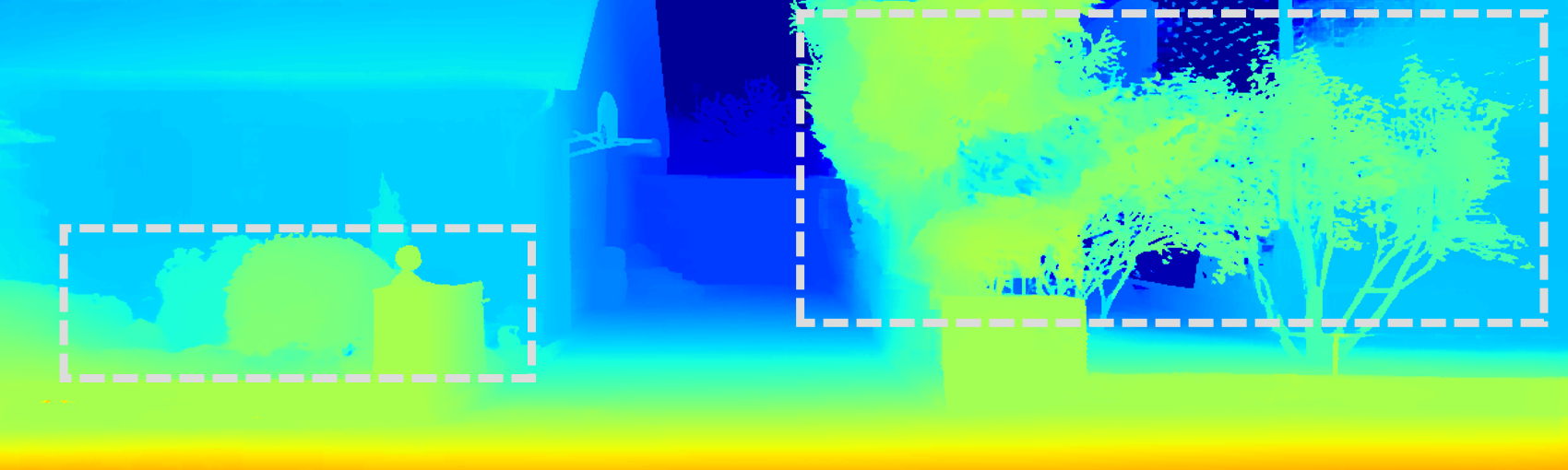}
        % \caption{StereoCarla}
    \end{subfigure}
    \begin{subfigure}[b]{0.13\textwidth}
        \centering
        \includegraphics[width=\textwidth]{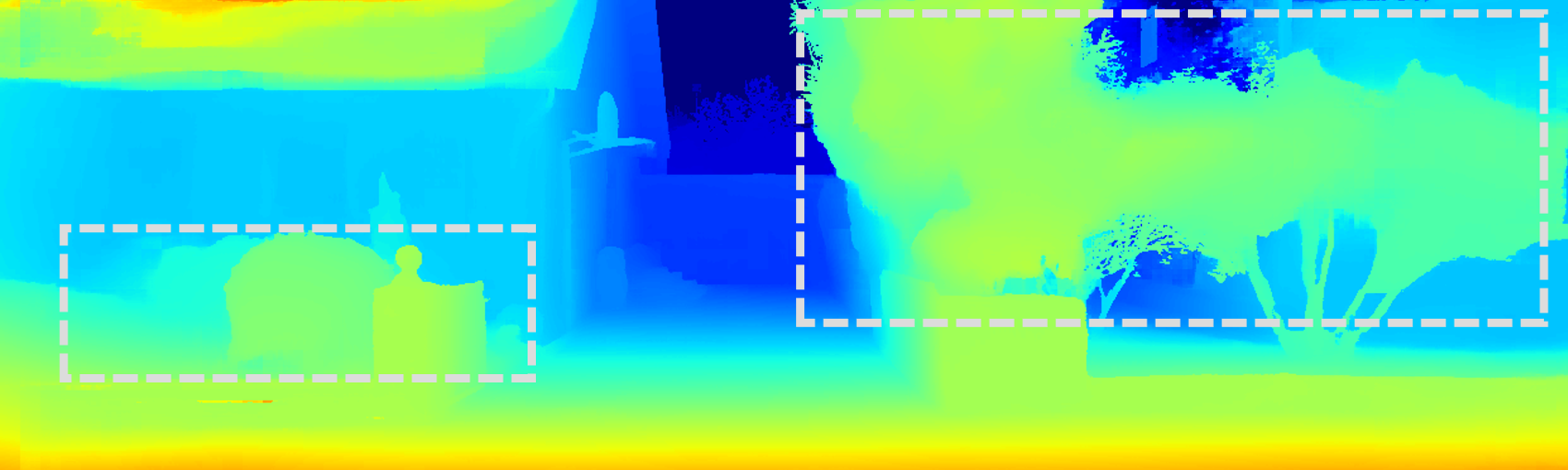}
        % \caption{Tartan}
    \end{subfigure}
    \begin{subfigure}[b]{0.13\textwidth}
        \centering
        \includegraphics[width=\textwidth]{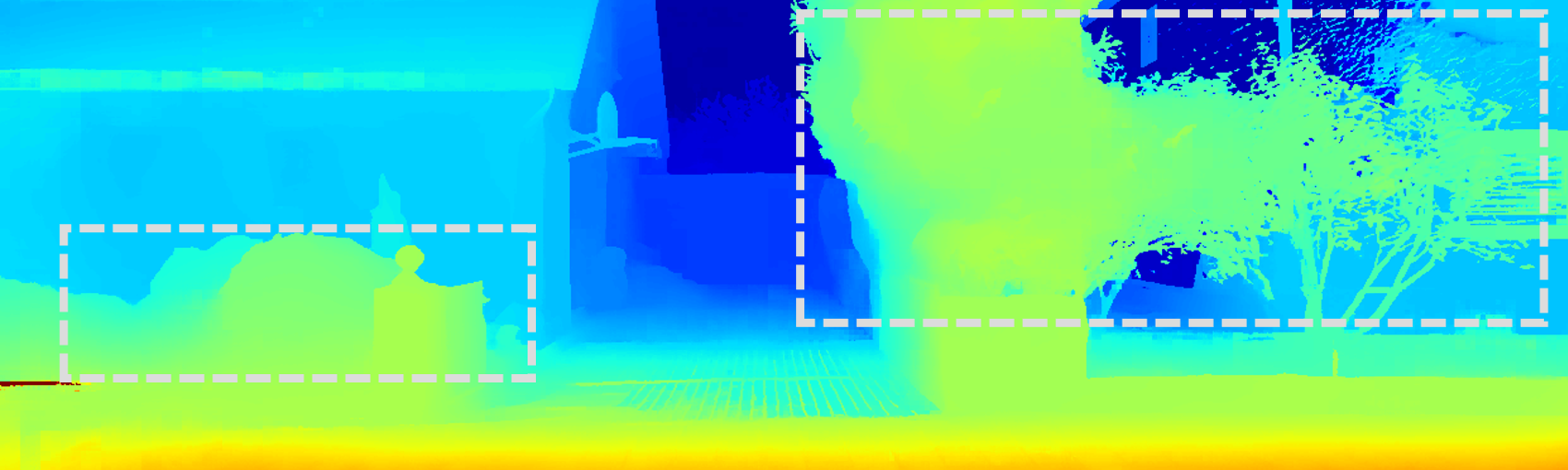}
        % \caption{CREStereo}
    \end{subfigure}
    \begin{subfigure}[b]{0.13\textwidth}
        \centering
        \includegraphics[width=\textwidth]{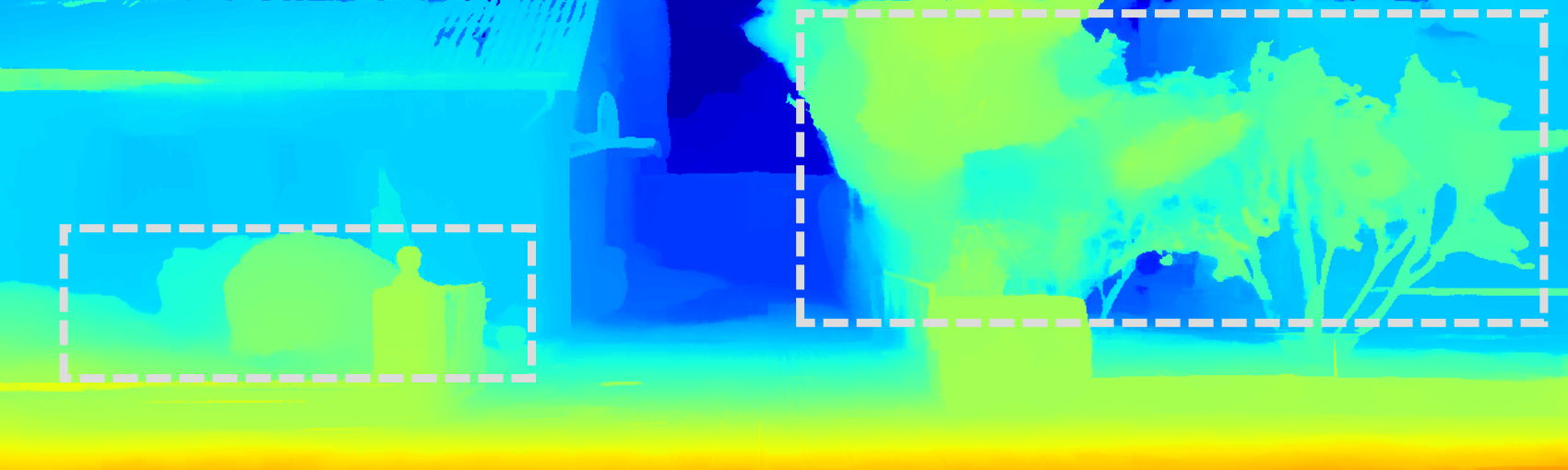}
        % \caption{Spring}
    \end{subfigure}
    \begin{subfigure}[b]{0.13\textwidth}
        \centering
        \includegraphics[width=\textwidth]{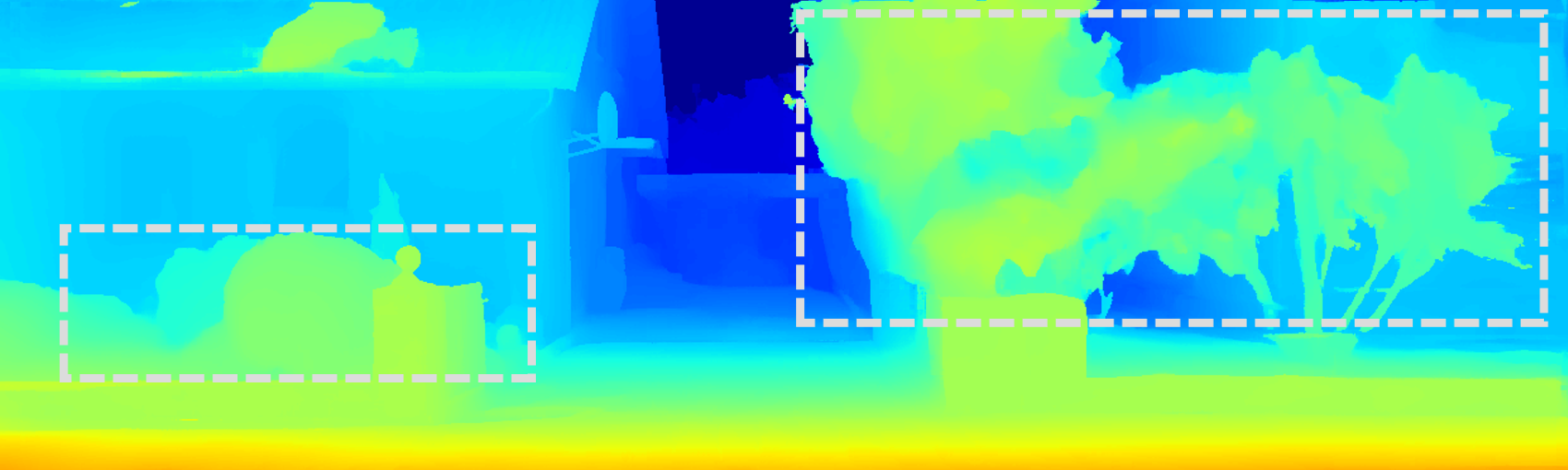}
        % \caption{Sintel}
    \end{subfigure}
    \begin{subfigure}[b]{0.13\textwidth}
        \centering
        \includegraphics[width=\textwidth]{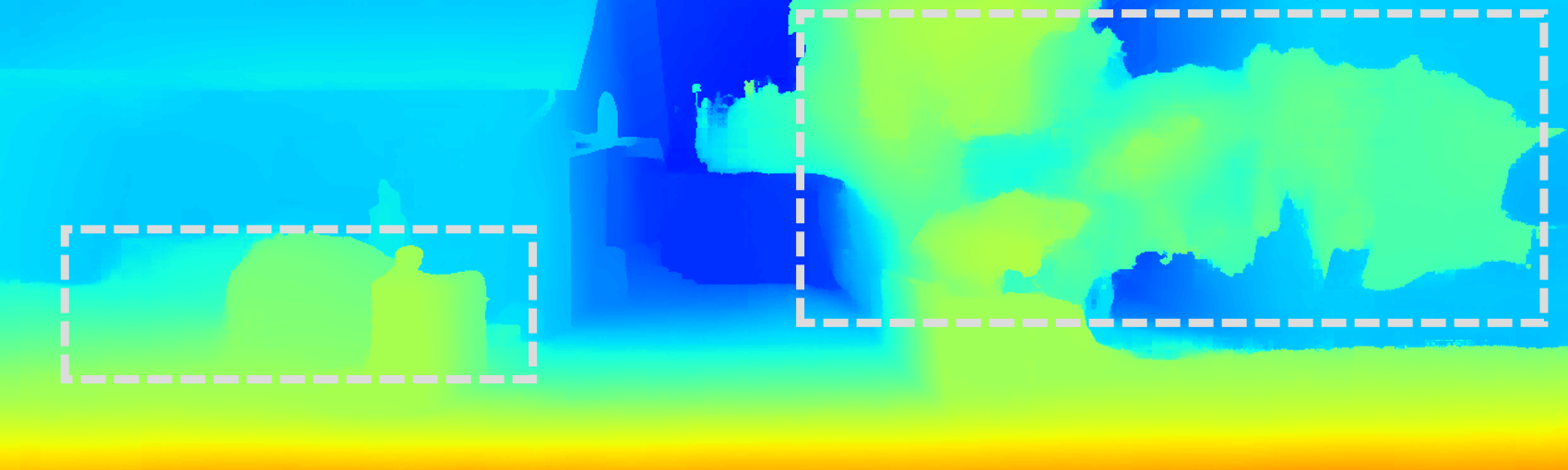}
        % \caption{Dynamic}
    \end{subfigure}
    % \begin{subfigure}[b]{0.13\textwidth}
    %     \centering
    %     \includegraphics[width=\textwidth]{figures/disp/k12_ft.png}
    %     % \caption{FT}
    % \end{subfigure}
    % \begin{subfigure}[b]{0.13\textwidth}
    %     \centering
    %     \includegraphics[width=\textwidth]{figures/disp/k12_instereo.png}
    %     % \caption{Instereo}
    % \end{subfigure}
    
%%%%%%%%%%%%%第二行
    \vspace{2mm}
    \begin{adjustbox}{rotate=90}\hspace{0.1mm}K15\end{adjustbox}
    \begin{subfigure}[b]{0.13\textwidth}
        \centering
        \includegraphics[width=\textwidth]{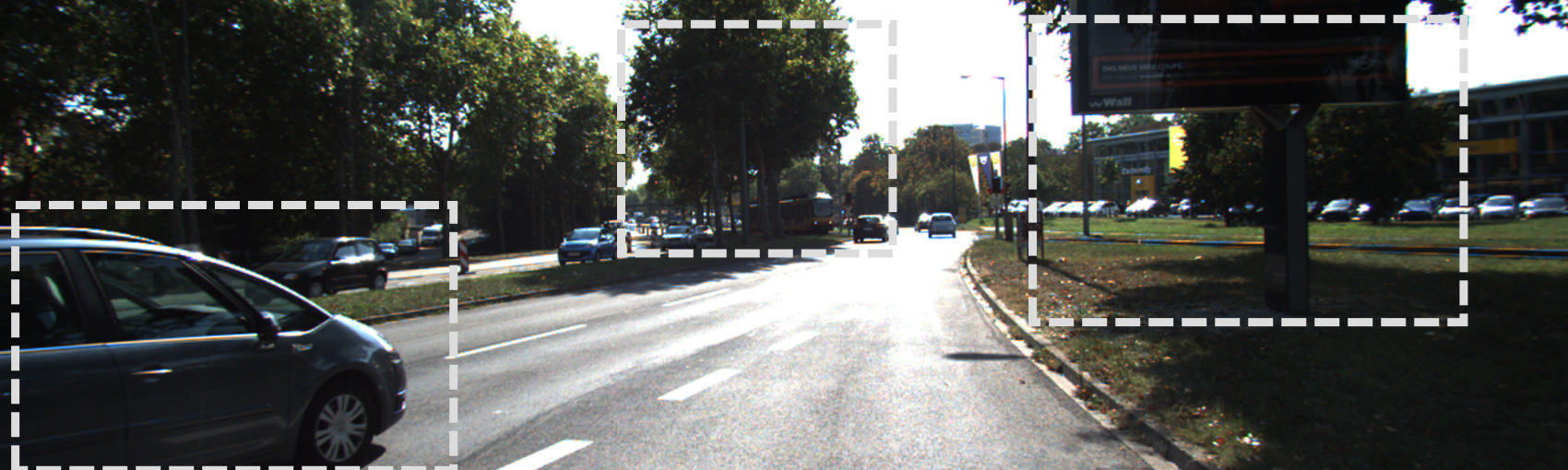}
        %\caption{RGB}
    \end{subfigure}
    \begin{subfigure}[b]{0.13\textwidth}
        \centering
        \includegraphics[width=\textwidth]{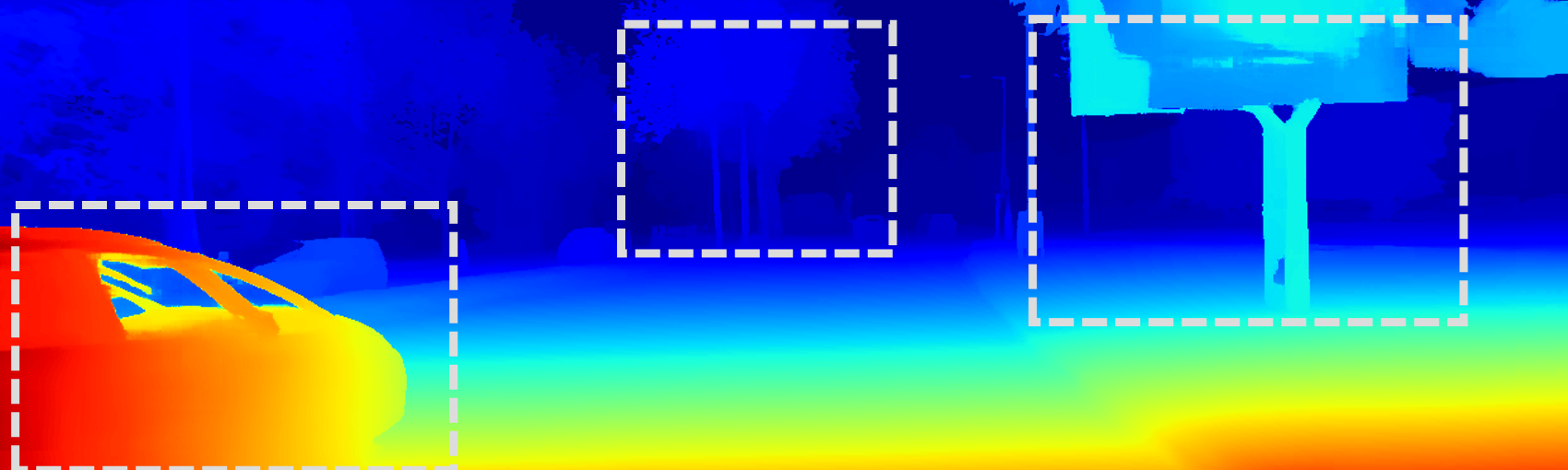}
        %\caption{StereoCarla}
    \end{subfigure}
    \begin{subfigure}[b]{0.13\textwidth}
        \centering
        \includegraphics[width=\textwidth]{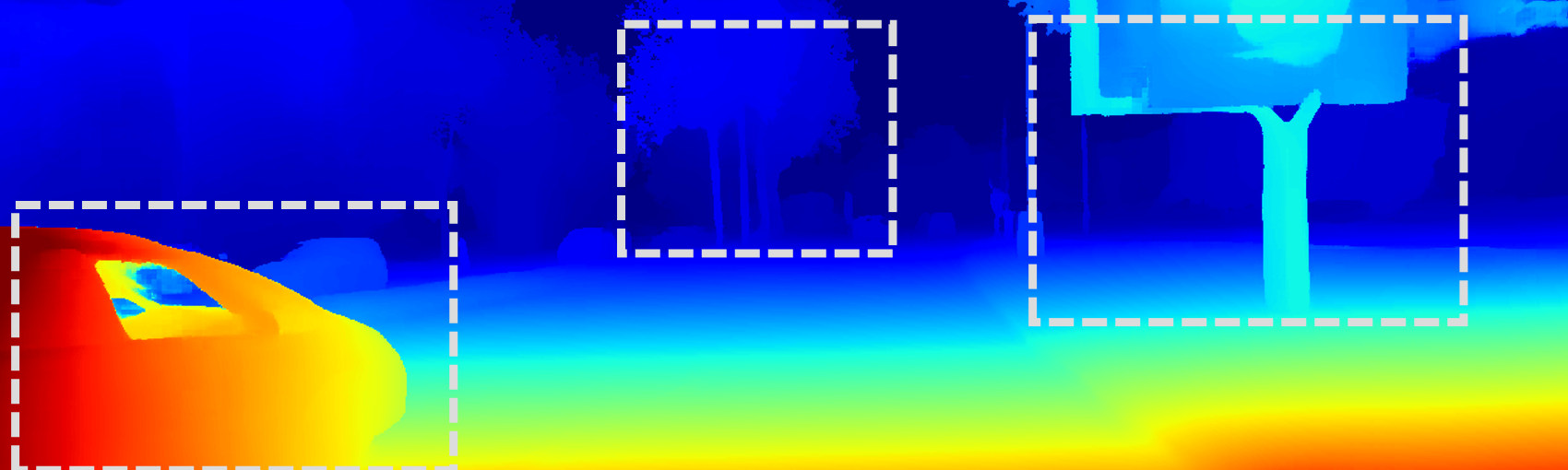}
        %\caption{Tartan}
    \end{subfigure}
    \begin{subfigure}[b]{0.13\textwidth}
        \centering
        \includegraphics[width=\textwidth]{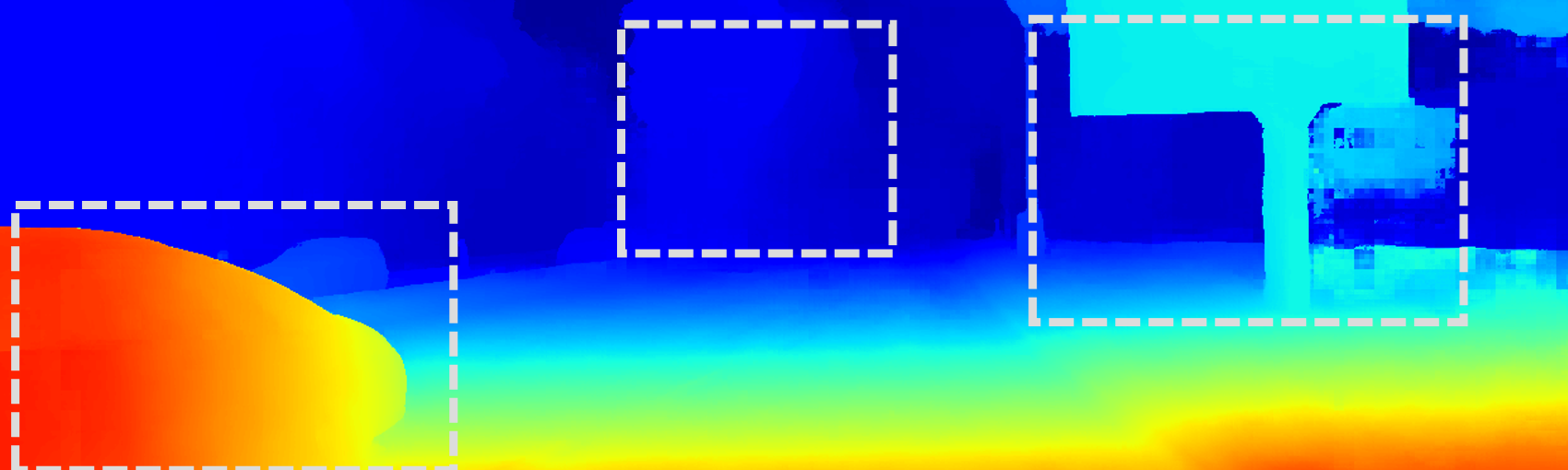}
        %\caption{CREStereo}
    \end{subfigure}
    \begin{subfigure}[b]{0.13\textwidth}
        \centering
        \includegraphics[width=\textwidth]{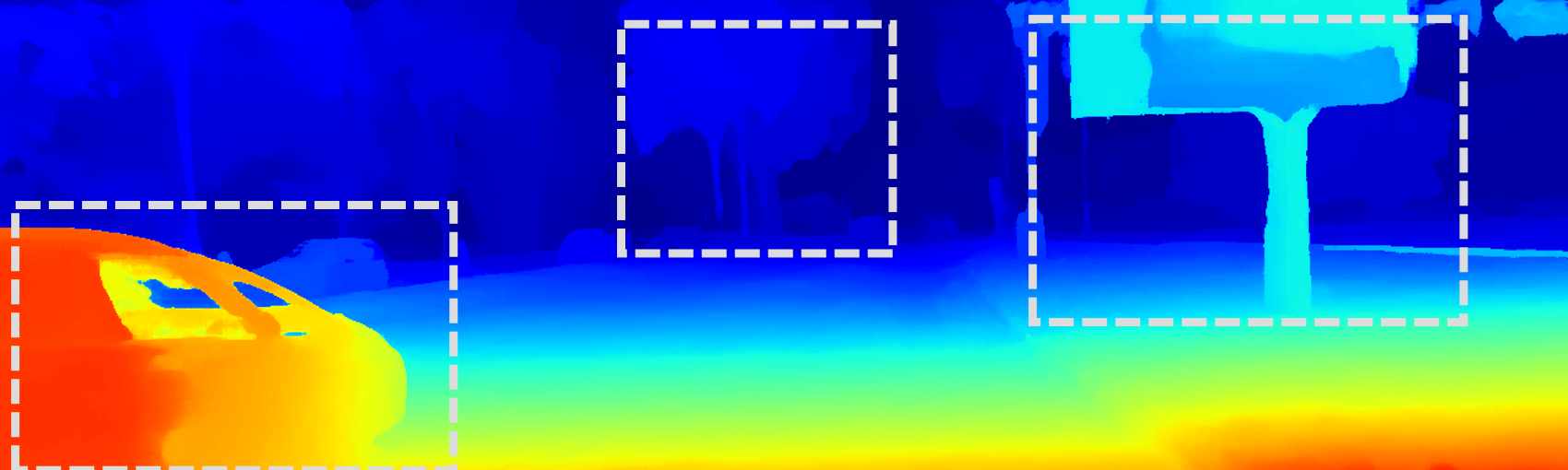}
        %\caption{Spring}
    \end{subfigure}
    \begin{subfigure}[b]{0.13\textwidth}
        \centering
        \includegraphics[width=\textwidth]{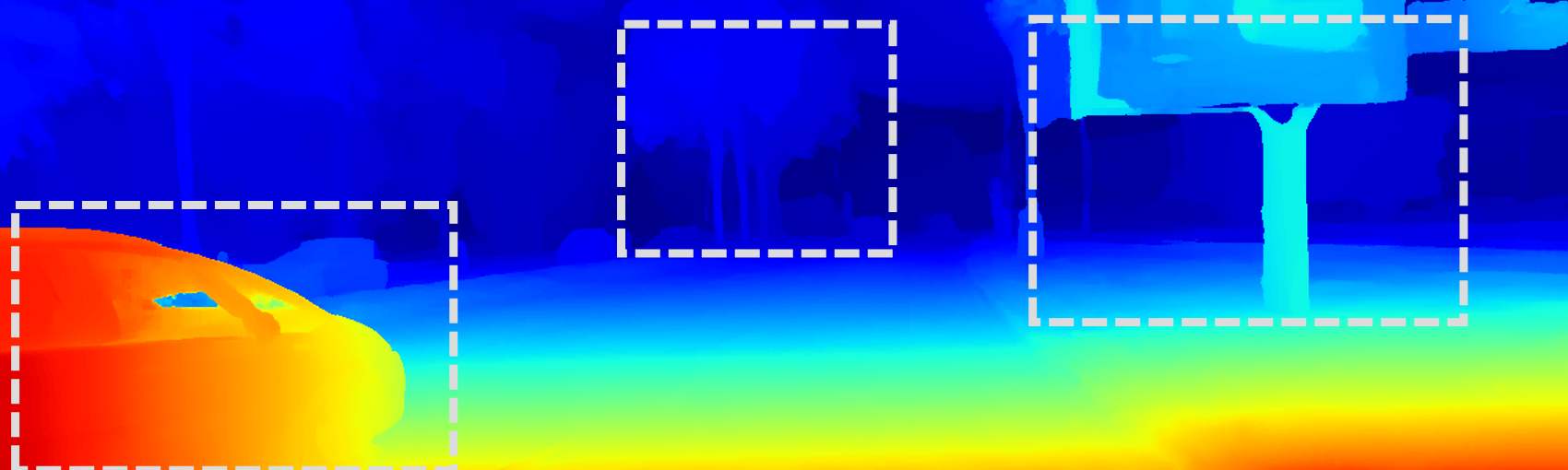}
        %\caption{Sintel}
    \end{subfigure}
    \begin{subfigure}[b]{0.13\textwidth}
        \centering
        \includegraphics[width=\textwidth]{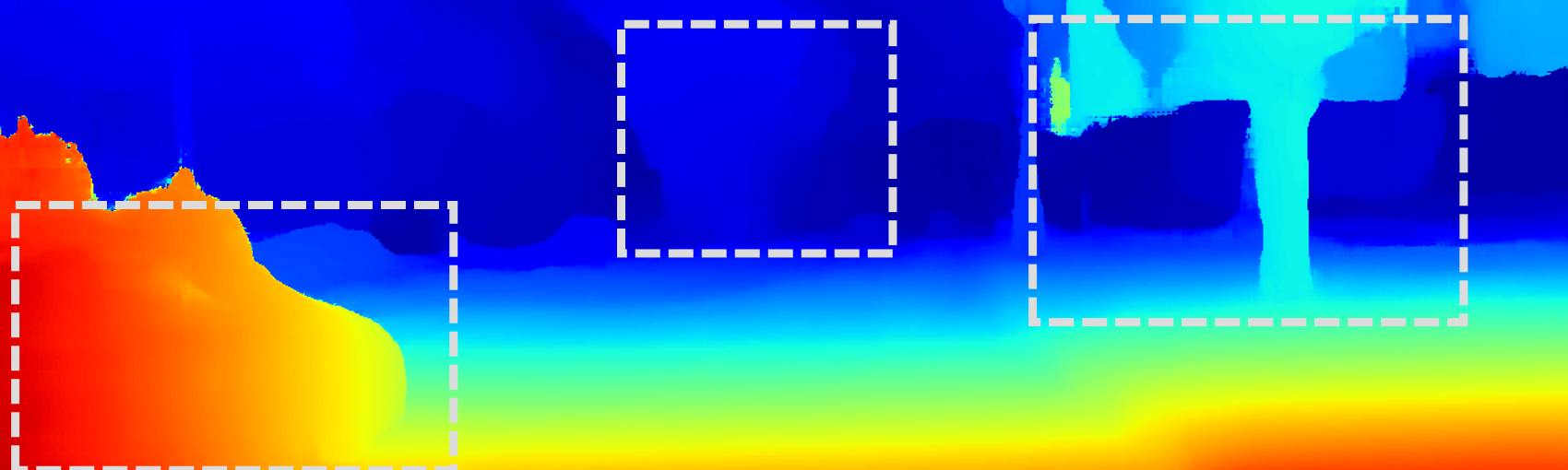}
        %\caption{Dynamic}
    \end{subfigure}
    % \begin{subfigure}[b]{0.13\textwidth}
    %     \centering
    %     \includegraphics[width=\textwidth]{figures/disp/k15_ft.png}
    %     %\caption{FT}
    % \end{subfigure}
    % \begin{subfigure}[b]{0.13\textwidth}
    %     \centering
    %     \includegraphics[width=\textwidth]{figures/disp/k15_instereo.png}
    %     %\caption{Instereo}
    % \end{subfigure}

%%%%%%%%%%%%%第三行
    \vspace{2mm}
    \begin{adjustbox}{rotate=90}\hspace{5mm}Midd\end{adjustbox}
    \begin{subfigure}[b]{0.13\textwidth}
        \centering
        \includegraphics[width=\textwidth]{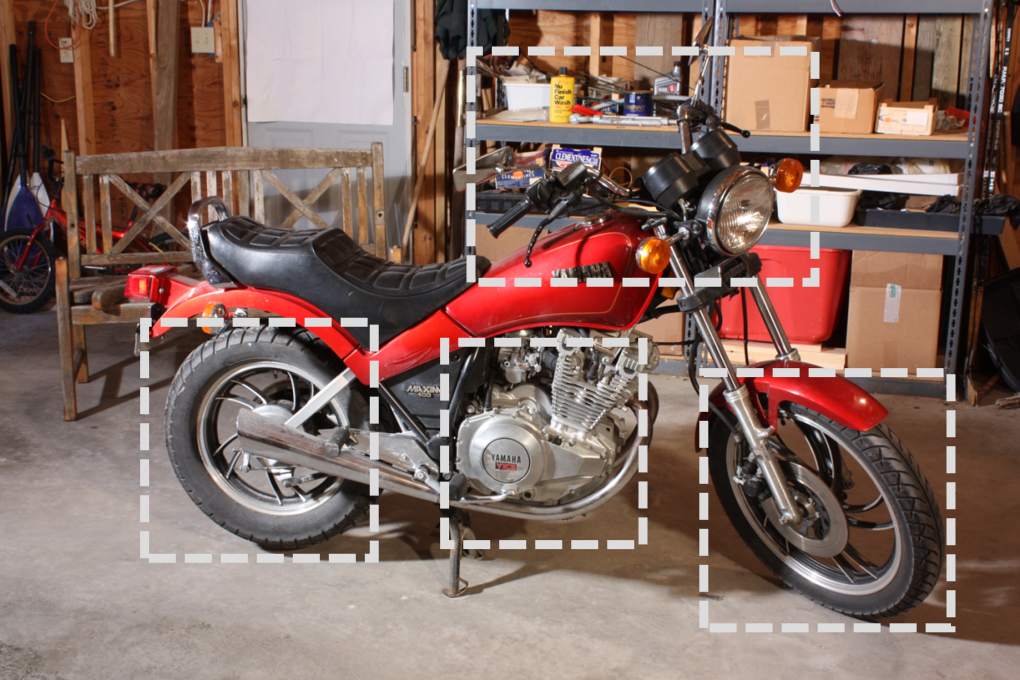}
        %\caption{RGB}
    \end{subfigure}
    \begin{subfigure}[b]{0.13\textwidth}
        \centering
        \includegraphics[width=\textwidth]{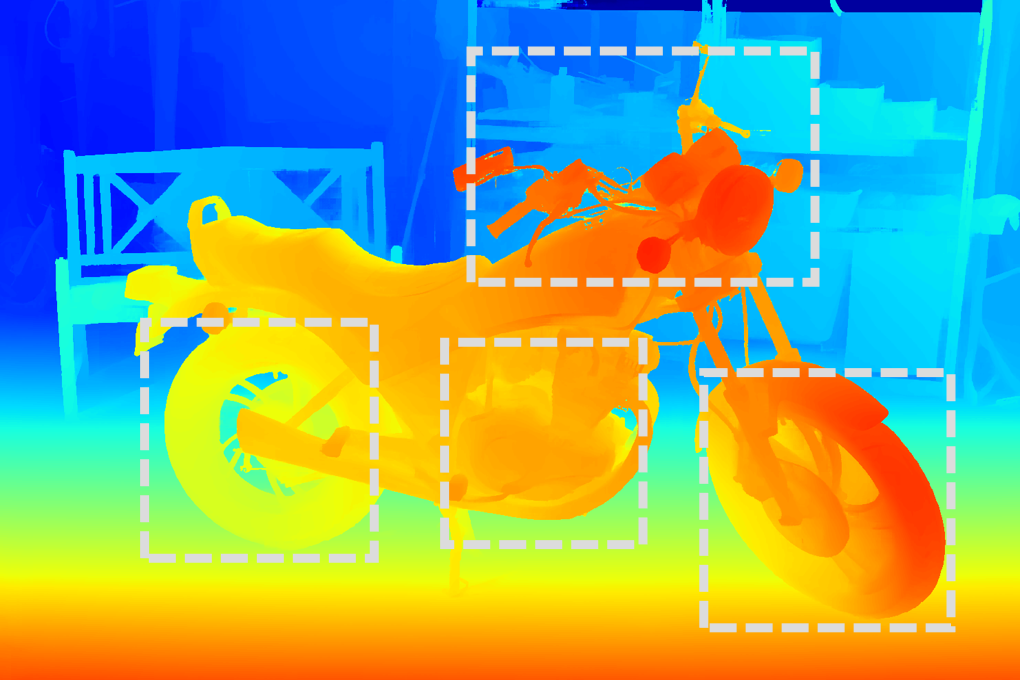}
        %\caption{StereoCarla}
    \end{subfigure}
    \begin{subfigure}[b]{0.13\textwidth}
        \centering
        \includegraphics[width=\textwidth]{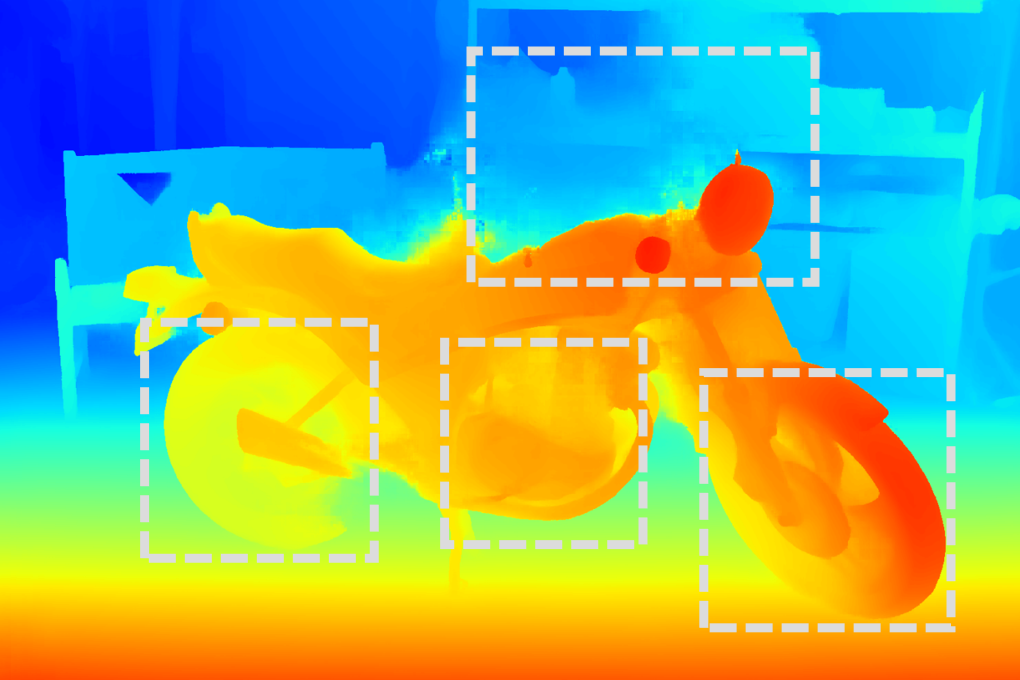}
        %\caption{Tartan}
    \end{subfigure}
    \begin{subfigure}[b]{0.13\textwidth}
        \centering
        \includegraphics[width=\textwidth]{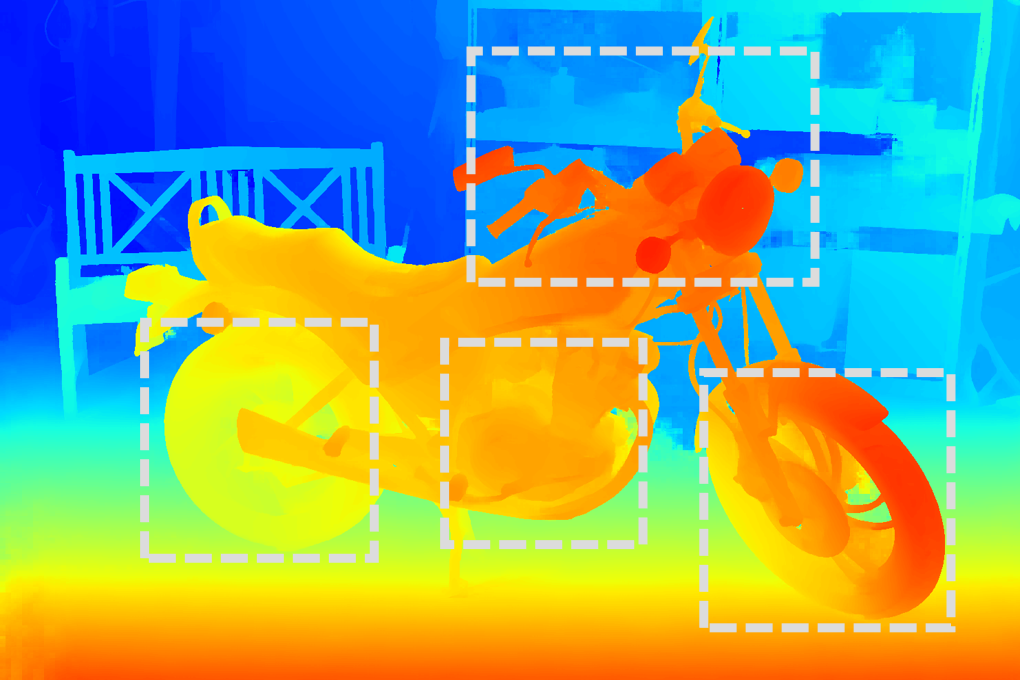}
        %\caption{CREStereo}
    \end{subfigure}
    \begin{subfigure}[b]{0.13\textwidth}
        \centering
        \includegraphics[width=\textwidth]{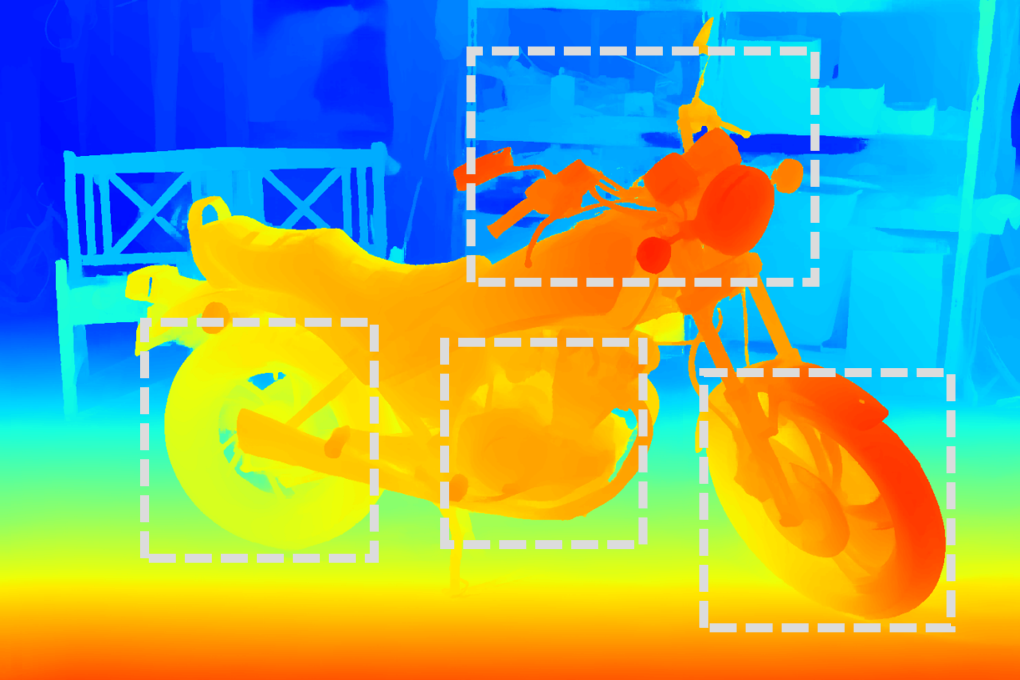}
        %\caption{Spring}
    \end{subfigure}
    \begin{subfigure}[b]{0.13\textwidth}
        \centering
        \includegraphics[width=\textwidth]{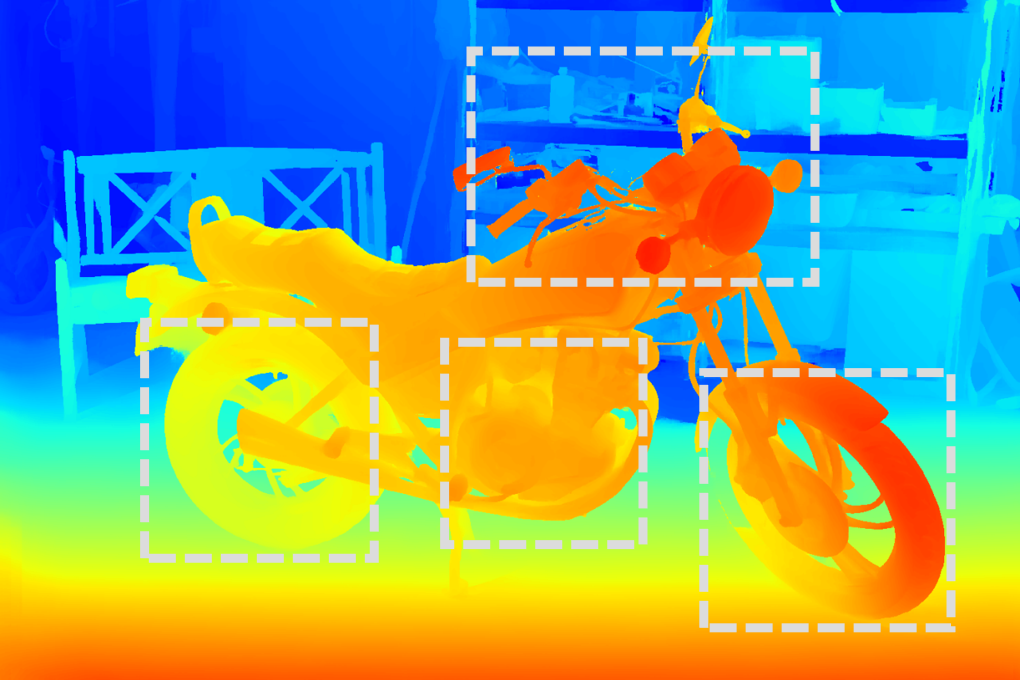}
        %\caption{Sintel}
    \end{subfigure}
    \begin{subfigure}[b]{0.13\textwidth}
        \centering
        \includegraphics[width=\textwidth]{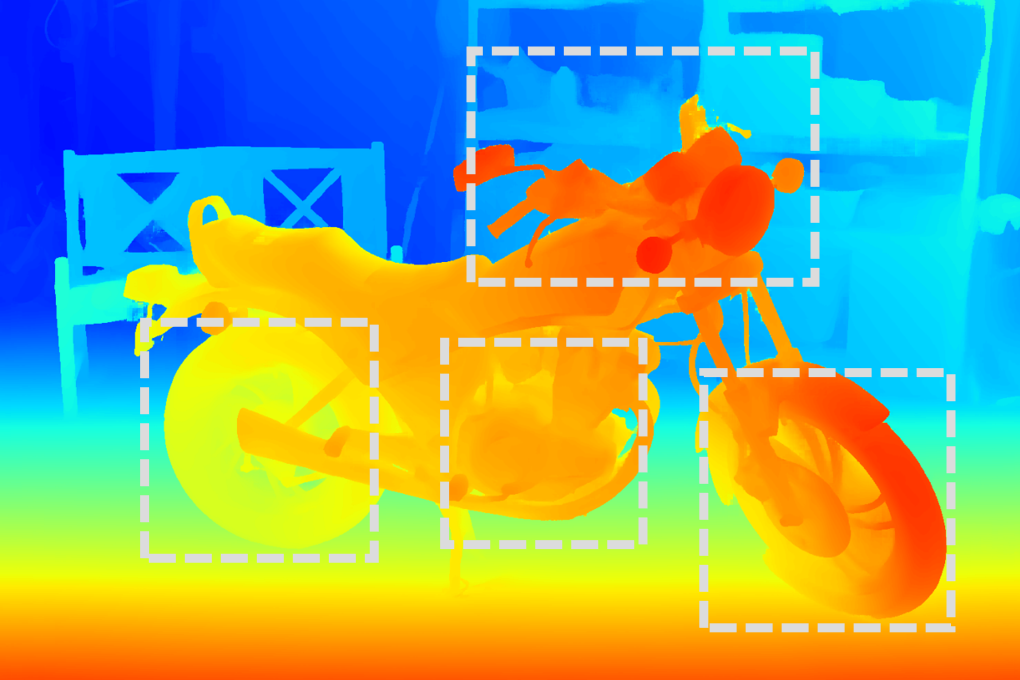}
        %\caption{Dynamic}
    \end{subfigure}
    % \begin{subfigure}[b]{0.13\textwidth}
    %     \centering
    %     \includegraphics[width=\textwidth]{figures/disp/mid_ft.png}
    %     %\caption{FT}
    % \end{subfigure}
    % \begin{subfigure}[b]{0.13\textwidth}
    %     \centering
    %     \includegraphics[width=\textwidth]{figures/disp/mid_instereo.png}
    %     %\caption{Instereo}
    % \end{subfigure}

%%%%%%%%%%%%%第4行
    \vspace{2mm}
    \begin{adjustbox}{rotate=90}\hspace{8mm}ETH3D\end{adjustbox}
    \begin{subfigure}[b]{0.13\textwidth}
        \centering
        \includegraphics[width=\textwidth]{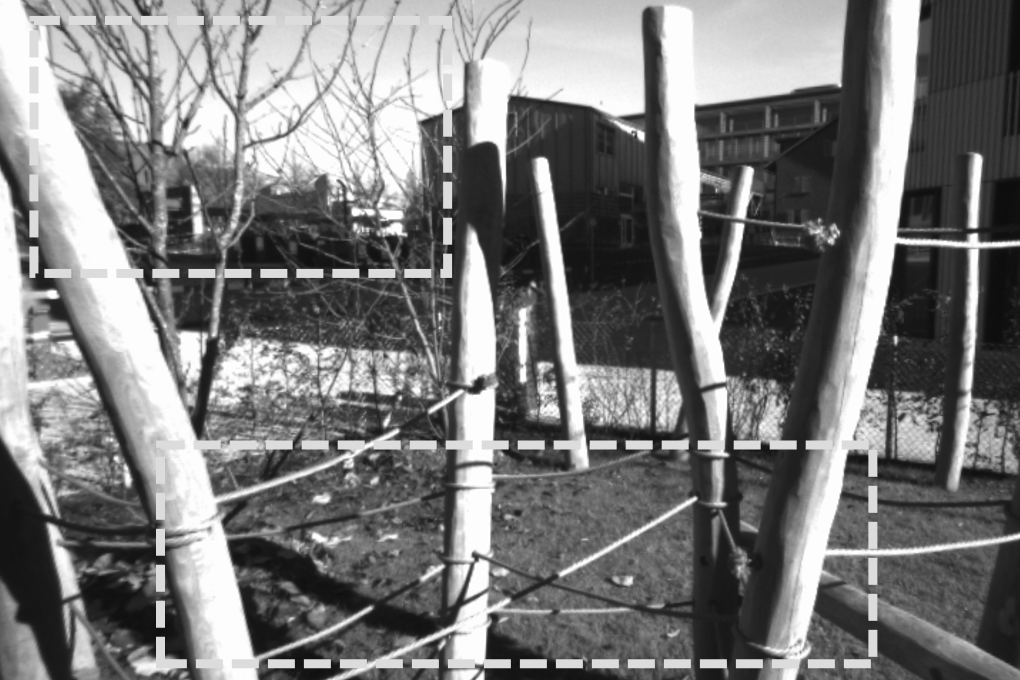}
        \caption{RGB}
    \end{subfigure}
    \begin{subfigure}[b]{0.13\textwidth}
        \centering
        \includegraphics[width=\textwidth]{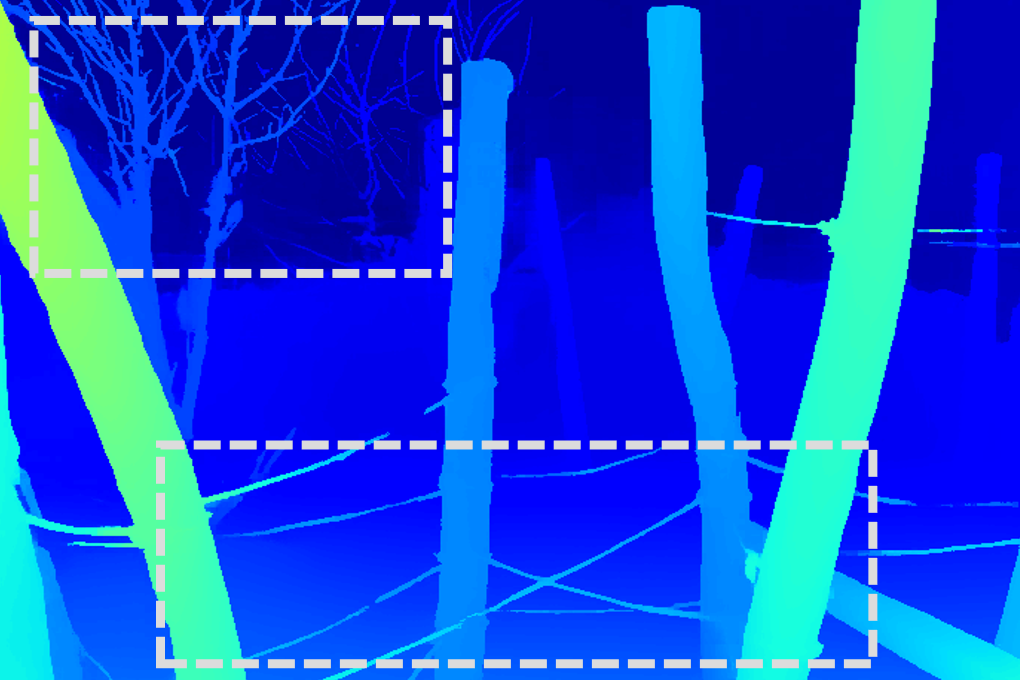}
        \caption{\textbf{SCarla (Ours)}}
    \end{subfigure}
    \begin{subfigure}[b]{0.13\textwidth}
        \centering
        \includegraphics[width=\textwidth]{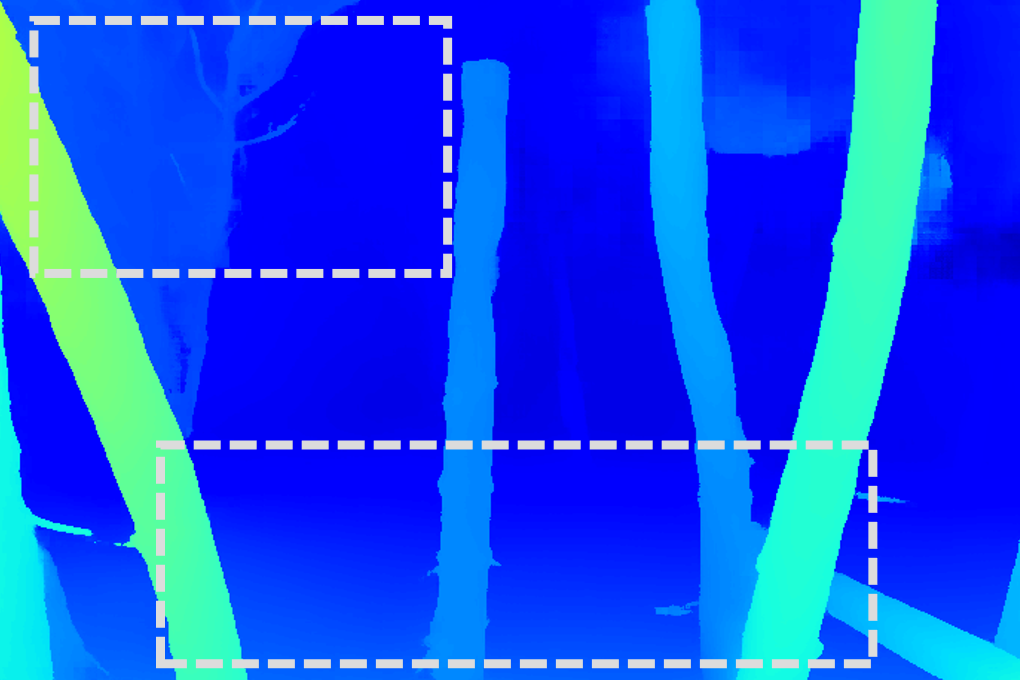}
        \caption{Tartan~\cite{wang2020tartanair}}
    \end{subfigure}
    \begin{subfigure}[b]{0.13\textwidth}
        \centering
        \includegraphics[width=\textwidth]{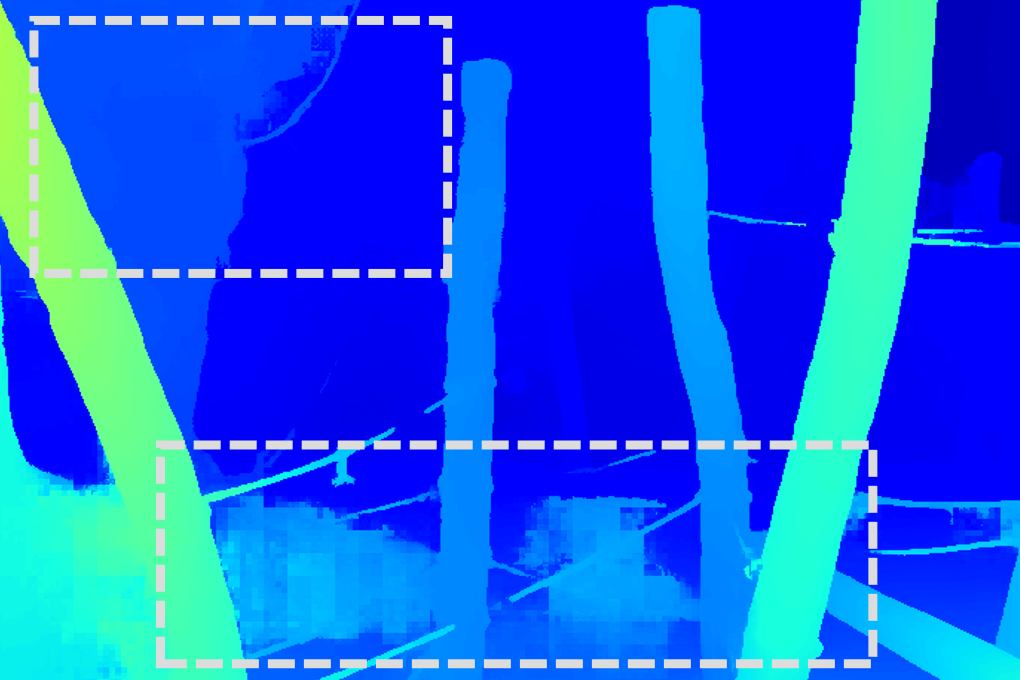}
        \caption{CRE~\cite{Crestereo}}
    \end{subfigure}
    \begin{subfigure}[b]{0.13\textwidth}
        \centering
        \includegraphics[width=\textwidth]{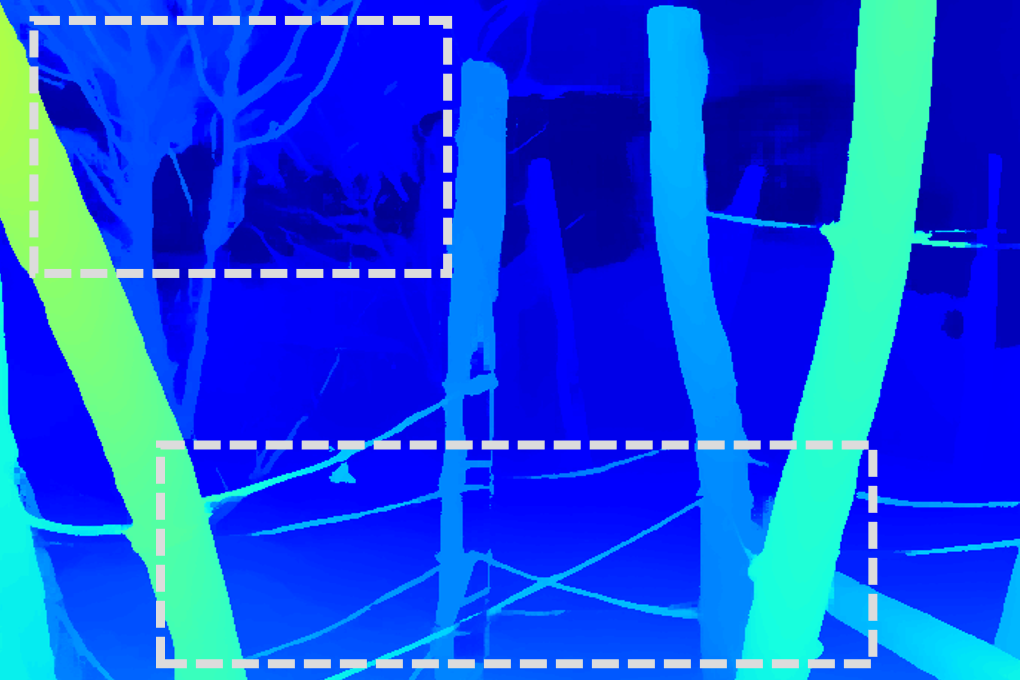}
        \caption{Spring~\cite{mehl2023spring}}
    \end{subfigure}
    \begin{subfigure}[b]{0.13\textwidth}
        \centering
        \includegraphics[width=\textwidth]{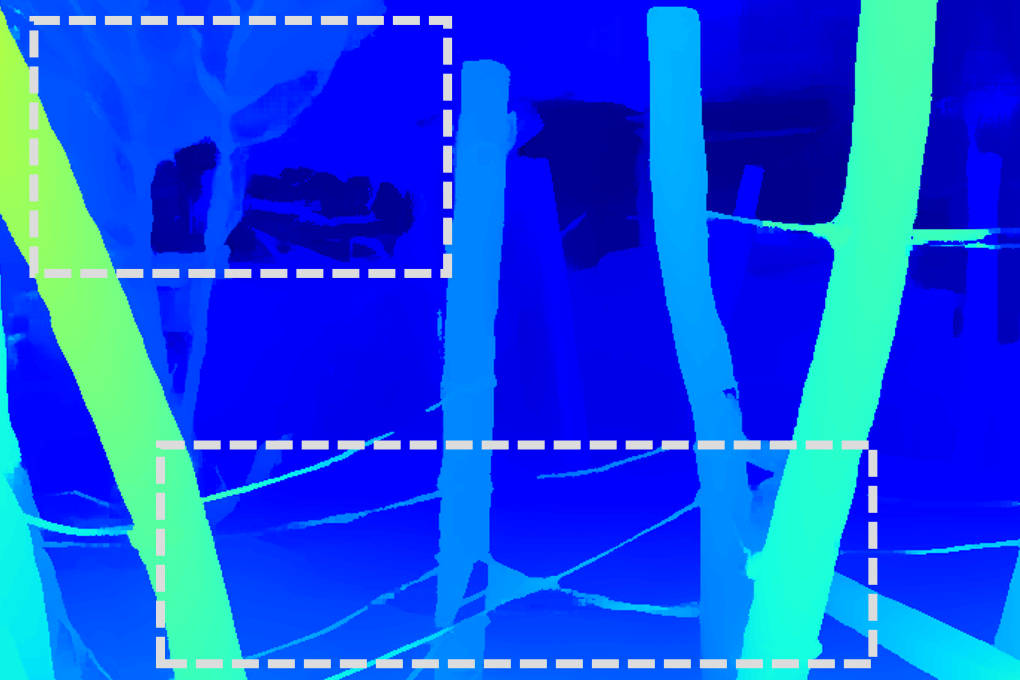}
        \caption{Sintel~\cite{Sintel}}
    \end{subfigure}
    \begin{subfigure}[b]{0.13\textwidth}
        \centering
        \includegraphics[width=\textwidth]{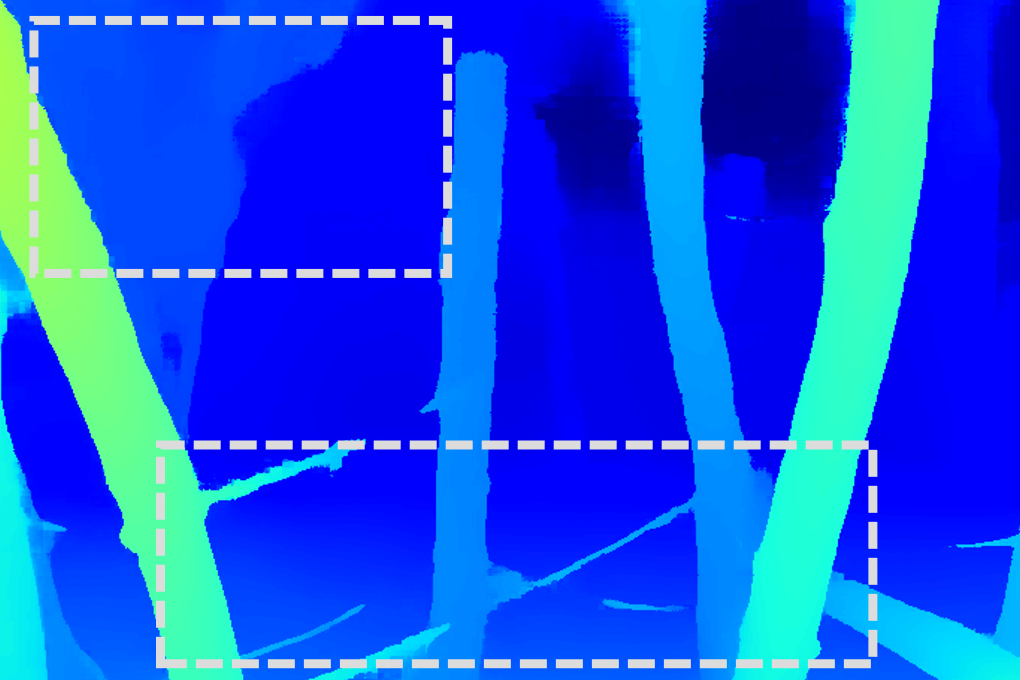}
        \caption{Dynamic~\cite{karaev2023dynamicstereo}}
    \end{subfigure}
    % \begin{subfigure}[b]{0.13\textwidth}
    %     \centering
    %     \includegraphics[width=\textwidth]{figures/disp/eth_ft.png}
    %     \caption{FT}
    % \end{subfigure}
    % \begin{subfigure}[b]{0.13\textwidth}
    %     \centering
    %     \includegraphics[width=\textwidth]{figures/disp/eth_instereo.png}
    %     \caption{Instereo}
    % \end{subfigure}

    \caption{\textbf{Qualitative comparison of models trained on different datasets.}}  % 总的Caption
    \label{fig:vis}
    % \vspace{-5mm}
\end{figure*}

\begin{table*}[t] 
\small
\centering
\caption{\textbf{Training on combinations of labeled stereo datasets.} Best results in \textbf{bold}.}%
% \scalebox{0.8}{
  % \ra{1.05}
  \setlength\tabcolsep{6pt}
  \renewcommand\arraystretch{1.1}
  % \scalebox{0.8}{
\begin{tabular}{l|ccccccccc||cccc|c}
  \thickhline
  \rowcolor{gray!20} 
\textbf{Mix}   & \textbf{SC} & \textbf{Tar}&  \textbf{CRE} & \textbf{SP}& \textbf{ST} &\textbf{DY}& \textbf{FT}&\textbf{I2K} &\textbf{VK2} &\textbf{Kitti12} & \textbf{Kitti15} & \textbf{Midd} & \textbf{Eth3D} & \textbf{Mean}  \\ %
  \hline\hline

  MIX 1 &\checkmark &&&&& &         &          &    & 4.11 & 4.87 & 9.12 & 3.17 & 5.32    \\%&     \\
  \rowcolor{gray!10}
  MIX 2 &\checkmark & \checkmark & &&&&&        &                    & 3.71 & 4.57 & 8.62 & 2.51 & 4.85\\%&     \\
  MIX 3 &\checkmark & \checkmark &   \checkmark   & &&&&&                & 3.86 & 4.48 & 8.14 & 2.26 & 4.69\\  %
  \rowcolor{gray!10}
  MIX 4 &\checkmark & \checkmark & \checkmark & \checkmark        &&&&&& 3.81 & 4.53 & 7.77 & \textbf{2.17} & 4.57\\
  MIX 5 & \checkmark & \checkmark &  \checkmark&\checkmark & \checkmark &&&&& 3.69 & 4.57 & 7.95 & 2.29 & 4.63\\%  \\
  \rowcolor{gray!10}
  MIX 6 & \checkmark & \checkmark & \checkmark &\checkmark & \checkmark& \checkmark &&&& 3.73 & 4.42 & 6.72 & 2.24 & 4.28\\%  \\
  MIX 7 & \checkmark & \checkmark & \checkmark &\checkmark & \checkmark & \checkmark& \checkmark&&& 3.71 & 4.46 & 6.81 & 2.50 & 4.37\\%  \\
  \rowcolor{gray!10}
  MIX 8 & \checkmark & \checkmark & \checkmark &\checkmark & \checkmark & \checkmark& \checkmark& \checkmark& &3.63 & 4.38 & 6.36 & 3.17 & 4.39 \\%  \\
  MIX 9 & \checkmark & \checkmark & \checkmark &\checkmark & \checkmark& \checkmark& \checkmark& \checkmark& \checkmark& \textbf{3.51} & \textbf{4.04} & \textbf{6.36} & 2.96 & \textbf{4.22} \\%
  % \midrule
  
  % ALL &\checkmark & \checkmark & \checkmark &\checkmark & \checkmark& \checkmark& \checkmark& \checkmark& \checkmark&\checkmark& \checkmark& 3.54 & \textbf{3.99} & 6.65 & 2.49 & \textbf{4.17} \\
  \thickhline
\end{tabular}
% }

\label{tab:dataset_mix_new}
\vspace{-3mm}
\end{table*}

\textbf{StereoCarla} For the in-domain experiments, towns 01–07 are used as the training set, while town 10 is reserved as the testing set, from which we select 3,500 stereo pairs. In contrast, all towns are utilized for training in cross-domain experiments.

\subsection{Evaluation metrics.}
For the in-domain experiments, the End-Point Error (EPE) is adopted for disparity evaluation, while Absolute Relative Error (Abs Rel) and threshold accuracy ($\delta < 1.25$) are used for depth evaluation to better appreciate the effect of the different baselines.
For cross-domain experiments, we assess our approach on four widely-used benchmarks: KITTI2012~\cite{kitti2012} (194 stereo pairs), KITTI2015~\cite{kitti2015} (200 pairs), Middlebury~\cite{middlebury} (15 image pairs, evaluated at half resolution), and ETH3D~\cite{eth3d} (27 grayscale stereo pairs). Following prior works~\cite{dsmnet2020,xu2023iterative}, we employ D1-all as the evaluation metric for KITTI, Bad 2.0 for Middlebury, and Bad 1.0 for ETH3D. D1-all reports the percentage of pixels whose disparity error exceeds 3 pixels. In contrast, Bad 1.0 and Bad 2.0 represent the proportion of pixels with errors larger than 1.0 and 2.0 pixels, respectively.

\begin{table}[t]
 % MG refers to Marigold~\cite{Marigold}.
\centering
\caption{\textbf{Zero-shot performance of the MIX datasets with and without StereoCarla}. Best results in \textbf{bold}.}
\label{tab:stereoCarla_ab_if}
\begin{tabular}{l|cccc|c}
\toprule
\rowcolor{gray!20} 
Setting & K12 & K15 & Midd & E3D  & Mean \\
\midrule
% SF & 8.67 & 7.46 & 16.36 & 23.46 & 13.99\\
% \midrule
MIX 7 (w/o SC) &3.94&\textbf{4.38}&8.57&13.04&7.48\\
\rowcolor{gray!10}
MIX 7 &\textbf{3.71} & 4.46 &\textbf{ 6.81} & \textbf{2.50} & \textbf{4.37}\\

\midrule
MIX 8(w/o SC) &3.64 &\textbf{4.19}&6.59&12.64& 6.77\\
\rowcolor{gray!10}
MIX 8 & \textbf{3.63} & 4.38 & \textbf{6.36} & \textbf{3.17} & \textbf{4.39}\\

\midrule
MIX 9 (w/o SC)&3.54 &\textbf{3.87}&6.72&9.59 &5.93\\
\rowcolor{gray!10}
MIX 9 &\textbf{3.51} & 4.04 & \textbf{6.36} & \textbf{2.96} & \textbf{4.22}\\
\bottomrule
\end{tabular}
\vspace{-5mm}
\end{table}

%BL
\begin{table*}[t]
\small
  \centering
\caption{\textbf{Ablation study of baseline on StereoCarla dataset.} Best results in \textbf{bold}.} 
\setlength\tabcolsep{5pt}
\renewcommand\arraystretch{1.0}
\scalebox{0.88}{
\begin{tabular}{l||c|c|c|c|c|c||cccc|c}
\thickhline

\rowcolor{gray!20}
 & \multicolumn{6}{c||}{\textbf{In-domain (StereoCarla)}} & \multicolumn{5}{c}{\textbf{Cross-domain}} \\
 
\rowcolor{gray!20}
& \multicolumn{1}{c|}{\textbf{BL(10)}}&
\multicolumn{1}{c|}{\textbf{BL(54)}}&
\multicolumn{1}{c|}{\textbf{BL(100)}}&
\multicolumn{1}{c|}{\textbf{BL(200)}}&
\multicolumn{1}{c|}{\textbf{BL(300)}}&
\multicolumn{1}{c||}{\textbf{Mean}}&
\textbf{K12} & \textbf{K15} & \textbf{Midd} & \textbf{E3D} &\\

\rowcolor{gray!20}
\multirow{-3}{*}{\textbf{Train\textbackslash Test}} 
& EPE$\downarrow$  
& EPE$\downarrow$    
& EPE$\downarrow$  
& EPE$\downarrow$  
& EPE$\downarrow$  
& EPE$\downarrow$  
& D1\_all$\downarrow$ & D1\_all$\downarrow$ & Bad2.0$\downarrow$ & Bad1.0$\downarrow$ & \multirow{-2}{*}{\textbf{Mean}} \\

\hline\hline
BL(10) & \underline{0.25} & 12.41 & 25.25 & 37.20 & 54.29 & 25.48 & 91.84 & 86.88 & 96.38 & 25.79 & 75.22 \\
\rowcolor{gray!10}
BL(54)   & 0.27 & \underline{1.15} & 1.63 & 4.07 & 7.26 & 2.88 &4.75 & 5.18 & 15.36 & \underline{3.69} & 7.25 \\
BL(100)  & 1.03 & 1.31 & \underline{1.19} & 3.09 & 5.30 & \underline{2.38} &\underline{4.17} & 5.13 & \underline{10.29} & 6.81 & \underline{6.60} \\
\rowcolor{gray!10}
BL(200)  & 0.27 & 1.30 & 1.62 & 3.97 & 7.52 & 2.94 
&4.25 & \textbf{4.87} & 10.90 & 23.04 & 10.77 \\
BL(300)  & 34.81 & 3.14 & 2.35 & \underline{2.80} & \textbf{3.64} & 9.75 & 4.82 & 4.98 & 14.57 & 41.11 & 16.37 \\
\rowcolor{gray!10}
\textbf{All} & \textbf{0.24} & \textbf{0.93} & \textbf{1.02} & \textbf{2.45} & \underline{3.79} & \textbf{1.69} &\textbf{4.11} & \textbf{4.87} & \textbf{9.12} & \textbf{3.17} & \textbf{5.32} \\

\thickhline
\end{tabular}
}
\label{tab:stereoCarla_ablation_BL}
% \vspace{-6.5mm}
\end{table*}

\subsection{Main Results}

To validate the effectiveness of our proposed dataset, we conduct extensive cross-domain evaluation by fine-tuning on various training sets and testing on four standard benchmarks: KITTI2012, KITTI2015, Middlebury, and ETH3D.

\subsubsection{Performance Comparison with Different Stereo Datasets}

Table~\ref{tab:onetraincolor} summarizes the cross-domain evaluation results of models fine-tuned on various stereo datasets, clearly demonstrating the effectiveness of our proposed StereoCarla dataset. Specifically, StereoCarla consistently achieves the best overall performance, surpassing existing datasets by a significant margin in terms of generalization accuracy. Compared to the model trained on SceneFlow, StereoCarla improves the mean error from 13.99 to 5.32, representing the largest overall enhancement among all evaluated datasets. It obtains the lowest disparity errors of 9.12 and 3.17 on Middlebury~\cite{middlebury} and ETH3D~\cite{eth3d} datasets respectively, significantly outperforming all other datasets evaluated. On KITTI2012~\cite{kitti2012} and KITTI2015~\cite{kitti2015} datasets, StereoCarla also ranks among the top-performing datasets, underscoring its strong generalization capability in complex outdoor scenes. In contrast, other datasets exhibit inconsistent performance across evaluation benchmarks, highlighting their limitations in robustness and generalizability. For instance, VirtualKITTI2~\cite{cabon2020virtual}, despite achieving excellent results on KITTI benchmarks, shows poor performance on ETH3D~\cite{eth3d}. Similarly, FallingThings demonstrates uneven accuracy, excelling in KITTI~\cite{kitti2012,kitti2015} datasets but underperforming significantly on ETH3D~\cite{eth3d}.

The experimental evidence strongly indicates that StereoCarla effectively bridges existing gaps in data diversity and domain coverage, offering a robust platform for training stereo matching models with outstanding generalization performance. As illustrated in Fig~\ref{fig:vis}, models trained on StereoCarla produce disparity maps with sharper object boundaries compared to those trained on other datasets.
These results demonstrate the strong cross-domain generalization and high-fidelity depth perception achieved by StereoCarla-trained models.

\subsubsection{Training on mixed stereo datasets}
Building upon the strong cross-domain results achieved by training on StereoCarla, we further explore its impact when used in combination with other labeled datasets. As shown in Table~\ref{tab:dataset_mix_new}, adding complementary datasets incrementally (MIX 2–MIX 9) continues to improve performance, confirming the compatibility and extensibility of StereoCarla with other sources. Notably, MIX 3 and MIX 4—constructed by combining StereoCarla with only 2–3 additional datasets—already surpass many larger combinations, highlighting the efficiency of StereoCarla in conveying diverse scene priors.
In the most comprehensive setup (MIX 9), which combines all datasets, we achieve the best overall performance across all benchmarks, with the lowest mean disparity error (4.22).
As shown in Table~\ref{tab:stereoCarla_ab_if}, the absence of StereoCarla leads to substantial performance degradation, particularly in complex environments like ETH3D and Middlebury, where the model's generalization capacity is severely limited without the dataset.
These results clearly indicate that StereoCarla not only excels as a standalone dataset but also acts as a critical backbone when scaling to multi-source training. Its high diversity and scene realism allow it to complement other datasets effectively, reinforcing the conclusion that StereoCarla is essential for building generalizable stereo matching systems.

\subsection{More analysis about StereoCarla datasets.}

\begin{table*}[t]
 % MG refers to Marigold~\cite{Marigold}.
\small
  \centering
\caption{\textbf{Ablation study of camera angles on StereoCarla dataset.} Best results in \textbf{bold}.}
\setlength\tabcolsep{5pt}
\renewcommand\arraystretch{1.1}
\scalebox{0.82}{
\begin{tabular}{l||cccccc|c||cccc|c}
\thickhline
\rowcolor{gray!20}
& \multicolumn{7}{c||}{\textbf{In-domain (StereoCarla)}} & \multicolumn{5}{c}{\textbf{Cross-domain}} \\

\rowcolor{gray!20}
\multirow{-2}{*}{\textbf{Train\textbackslash Test}} & \multicolumn{1}{c|}{\textbf{Normal}}&
\multicolumn{1}{c|}{\textbf{Roll (5)}}&
\multicolumn{1}{c|}{\textbf{Roll (15)}}&
\multicolumn{1}{c|}{\textbf{Roll (30)}}&
\multicolumn{1}{c|}{\textbf{Pitch (0)}}&
\multicolumn{1}{c|}{\textbf{Pitch (-30)}}&
\multicolumn{1}{c||}{\textbf{Mean}}&
\textbf{K12} & \textbf{K15} & \textbf{Midd} & \textbf{E3D} &\textbf{Mean}  \\

\hline\hline
Normal & \underline{1.83} & 1.74 & 1.17 & 5.84 & 0.97 & 0.59 & \underline{1.86} & \underline{4.21} & \underline{5.15} & 14.2 & 3.59 & 6.79 \\
\rowcolor{gray!10}
Roll(5) & 1.97 & \underline{1.62} & 1.09 & 5.58 & 1.00 & 0.58 & 1.97 & 4.40 & 5.24 & \underline{11.78} & \textbf{3.09} & \underline{6.13} \\
Roll(15) & 2.28 & 2.07 & \textbf{0.99} & 4.61 & 1.06 & 0.62 & 1.94 & 5.24 & 5.51 & 12.45 & 3.84 & 6.76 \\
\rowcolor{gray!10}
Roll(30) & 2.67 & 2.49 & 1.38 & \textbf{2.88} & 1.10 & 0.69 & 1.87 & 6.05 & 5.90 & 14.46 & 4.07 & 7.62 \\
Pitch(0) & 2.77 & 2.19 & 1.77 & 7.02 & \underline{0.90} & 0.68 & 2.39 & 5.02 & 5.48 & 13.65 & 3.54 & 6.92 \\
\rowcolor{gray!10}
Pitch(-30) & 7.89 & 4.04 & 4.49 & 10.89 & 1.53 & \textbf{0.51} & 4.89 & 20.29 & 18.97 & 17.04 & 5.51 & 15.45 \\
\textbf{All} & \textbf{1.74} & \textbf{1.60} & \underline{1.00} & \underline{3.04} & \textbf{0.87} & \underline{0.52} & \textbf{1.46} & \textbf{4.11} & \textbf{4.87} & \textbf{9.12} & \underline{3.17} & \textbf{5.32} \\

\thickhline
\end{tabular}
}
\label{tab:stereoCarla_ablation_camera}
% \vspace{-6.5mm}
\end{table*}

\begin{table}[t]
\small
  \centering
\caption{\textbf{Ablation study on the weather on the StereoCarla dataset.} Best results in \textbf{bold}.}
\setlength\tabcolsep{5pt}
\renewcommand\arraystretch{1.1}
\scalebox{0.8}{
\begin{tabular}{l||cccc|c}
\thickhline
\rowcolor{gray!20}
\textbf{Setting} &  \textbf{Kitti12} & \textbf{Kitti15} & \textbf{Middlebury} & \textbf{Eth3D}& \textbf{Mean} \\
\hline\hline
w/o Wea& 4.15&5.04&9.61& \textbf{2.99} &5.45\\
\rowcolor{gray!10} 
\textbf{All} & \textbf{4.11}& 	\textbf{4.87}& 	\textbf{9.12}	& 3.17	& \textbf{5.32} \\
\thickhline
\end{tabular}}
\label{tab:stereoCarla_ablation}
\vspace{-6.5mm}
\end{table}

We also analyze the impact of varying baseline distances and camera viewing angles on the performance of stereo-matching models trained on the StereoCarla dataset. 

\textbf{Baseline} The results presented in Table~\ref{tab:stereoCarla_ablation_BL} reveal several key insights:
In the in-domain evaluation, extremely short baselines such as 10 cm lead to highly accurate disparity estimation within the same baseline configuration (EPE = 0.25) but cause a dramatic degradation when tested on larger baselines (e.g., EPE = 54.29 at 300 cm), indicating severe overfitting to the training geometry. Conversely, moderate baselines such as 54 cm, 100 cm, and 200 cm demonstrate a more balanced performance across baseline changes, with BL(100) achieving the second-best mean EPE of 2.38 and stable depth accuracy, highlighting its adaptability to geometric variation. Large baselines like 300 cm achieve competitive accuracy when tested on similar scales (EPE = 3.64) but generalize poorly to shorter baselines, suggesting a scale-specific bias in learned disparity features. The “All” configuration, which jointly trains on data from all baselines, consistently achieves the best overall performance both in-domain (mean EPE = 1.69) and cross-domain, with the lowest errors on KITTI 2012, KITTI 2015, Middlebury, and ETH3D benchmarks, underscoring the importance of baseline diversity for enhancing both robustness and generalization.

\textbf{Camera angles} As shown in Table~\ref{tab:stereoCarla_ablation_camera}, models trained on single-angle datasets generally perform best when tested on the same or similar camera orientations, but their performance degrades on other viewpoints. For instance, the model trained with a 30° roll angle achieves the lowest EPE (2.88) on test data with the same 30° roll, but shows higher errors on other angles, indicating limited viewpoint generalization. Similarly, the model trained with a -30° pitch exhibits notably worse in-domain mean EPE (4.89) and significantly degraded cross-domain performance, reflecting the difficulty of extreme pitch variations.
In contrast, training on the combined dataset incorporating all camera angles ("All") consistently yields the best average performance both in-domain and across all cross-domain benchmarks. Specifically, this model achieves the lowest mean EPE of 1.46 in-domain and 5.32 on cross-domain datasets, outperforming any single-angle trained model by a significant margin. This result strongly supports the hypothesis that incorporating diverse camera angles during training substantially enhances the model's robustness and cross-view generalization capability.
Overall, these findings highlight the critical importance of camera viewpoint diversity in training data for developing stereo matching algorithms capable of handling the wide variety of camera poses encountered in real-world autonomous driving scenarios.

\textbf{Weather Conditions} We also compare stereo models trained with and without weather variations on the StereoCarla dataset. As shown in Table \ref{tab:stereoCarla_ablation}, including diverse weather conditions such as rain and fog also leads to consistent improvements on KITTI and Middlebury benchmarks. The mean EPE decreases from 5.45 to 5.32 across datasets, demonstrating enhanced robustness. This confirms that weather diversity is also important for generalizable stereo matching in autonomous driving.

\section{Conclusion}

In this work, we introduced \textit{StereoCarla}, a high-fidelity synthetic driving dataset designed to advance generalizable stereo matching. Built upon the CARLA simulator, \textit{StereoCarla} offers extensive diversity in camera baselines, viewpoints, and scene contexts—including challenging weather and illumination—thereby addressing the limited coverage of existing stereo datasets. Through comprehensive cross-domain experiments and ablation studies, we demonstrated that StereoCarla consistently improves the generalization capability of stereo models, outperforming 11 widely used datasets across multiple benchmarks. Our results further emphasize the critical role of dataset diversity, showing that both geometric factors (e.g., baseline and viewpoint) and semantic factors (e.g., scene and weather) are essential for robust stereo matching. We envision \textit{StereoCarla} as a scalable and extensible resource that can serve as a strong foundation for future research in generalizable stereo matching.

\textbf{Acknowledgements.} This work was supported by the National Natural Science Foundation of China under Grant 62373356 and the Joint Funds of the National Natural Science Foundation of China under U24B20162.

%%%%%%%%% REFERENCES

\bibliographystyle{IEEEtran}
\bibliography{main}

\end{document}